\documentclass[letterpaper, 10 pt, conference]{ieeeconf}

\IEEEoverridecommandlockouts
\usepackage[utf8]{inputenc}
\usepackage[T1]{fontenc}

\usepackage{graphicx}
\usepackage{amsmath}
\usepackage{url}
\usepackage{amssymb}
\usepackage{colortbl}
\usepackage{xcolor}
\usepackage{booktabs}
\usepackage[ruled, linesnumbered]{algorithm2e}
\usepackage[font=footnotesize]{caption}
\usepackage{multirow}
\usepackage{makecell}

\title{\LARGE \bf
Safe and Stylized Trajectory Planning for Autonomous Driving via Diffusion Model}

\author{Shuo Pei$^{1}$, Qin Li$^{2}$, Yong Wang$^{1,*}$, Yuanchen Zhu$^{2}$, Chen Sun$^{1}$, Yanan Zhao$^{2}$, Huachun Tan$^{2}$
\thanks{*This work was supported by the National Key Research and Development Program of China (Grant No. 2023YFB2504704-02), the Shandong Provincial Key Research and Development Program (Grant No. 2023CXPT032), and the Research Grants Council of Hong Kong (Grant No. 27206525).}
\thanks{$^{1}$Shuo Pei, Chen Sun, and Yong Wang (Corresponding author) are with the Department of Data and Systems Engineering, The University of Hong Kong, Hong Kong, China. (e-mail: wy0304@hku.hk).
$^{2}$Yuanchen Zhu, Qin Li, Yanan Zhao, Huachun Tan are with the Beijing Institute of Technology, Beijing, China.}}

\begin{document}

\maketitle
\thispagestyle{empty}
\pagestyle{empty}

\begin{abstract}
Achieving safe and stylized trajectory planning in complex real-world scenarios remains a critical challenge for autonomous driving systems. This paper proposes the \textit{SDD Planner}, a diffusion-based framework that explicitly reconciles safety constraints with personalized driving styles in real time. (1) A Multi-Source Style-Aware Encoder that uses distance-sensitive attention to fuse dynamic agents and scene context, enabling joint perception of safety and driving style. (2) A Style-Guided Dynamic Trajectory Generator that applies adaptive safety--style guidance during the diffusion denoising process and integrates a feasibility-oriented post-processing stage to produce user-preferred and physically executable trajectories. Extensive experiments demonstrate the effectiveness of SDD Planner. On the StyleDrive benchmark, it improves the SM-PDMS metric by 3.9\% over the strongest baseline. On the NuPlan Test14 and Test14-hard benchmarks, SDD Planner ranks first with overall scores of 91.76 and 80.32, respectively, outperforming strong planning baselines such as PLUTO. Additional robustness and ablation studies further confirm the effectiveness of the proposed hyper-parameter design, learnable distance-sensitive attention, and cost-based trajectory selection. A deployment-oriented study with a distilled variant on an NVIDIA Orin platform, accelerated by DDIM fast sampling and mixed-precision execution, achieves an end-to-end latency of approximately $93\,\text{ms}$, indicating the practical executability of the proposed planner under onboard runtime constraints.
\end{abstract}

\begin{keywords}
Autonomous driving, trajectory planning, diffusion models, multi-modal encoding, stylized driving
\end{keywords}

\section{INTRODUCTION}
\label{sec:introduction}

Autonomous driving (AD) has emerged as a transformative technology in modern transportation, aiming to enhance road safety, traffic efficiency, and user comfort through intelligent perception, decision-making, and control \cite{tang2023multi, tang2025ftp, wang2025modelfreecontrol}. Central to an AD system is its ability to perceive the surrounding environment, predict the behaviors of nearby agents, and plan safe, efficient, and human-like trajectories under complex and uncertain traffic conditions. Among these modules, trajectory planning serves as the core component that directly bridges high-level behavioral reasoning with low-level vehicle control, fundamentally determining both the safety and ride experience of autonomous vehicles. Traditional rule-based planning methods generate trajectories based on predefined traffic rules and handcrafted heuristics, demonstrating high stability and interpretability in structured environments~\cite{dauner2023pdm}. However, their dependence on manually designed rules limits adaptability to unstructured or interactive scenarios, and updating these rules requires extensive engineering effort~\cite{dauner2023pdm, caesar2023nuplan}.

In contrast, learning-based planners have gained attention for their flexibility and capacity to model complex driving behaviors from data. Particularly, end-to-end learning paradigms have shown promising results in both simulation and real-world deployments, enabling adaptive and context-aware trajectory generation across diverse traffic scenarios~\cite{caesar2023nuplan}. Among these approaches, diffusion-based planning has recently emerged as a powerful generative modeling technique capable of capturing multimodal trajectory distributions through iterative denoising processes~\cite{sohl2015deep, chi2023diffusion}. This capability allows for generating diverse and physically plausible driving behaviors, offering advantages over conventional learning-based methods that often converge to single-modal or overly conservative solutions~\cite{janner2022diffuser, zheng2025diffusion}. Despite these advancements, current diffusion-based planners remain limited in user-centric deployment. Existing methods often employ fixed safety--style weights, lack dynamic adaptation to heterogeneous risks, and insufficiently model interactions with dynamic agents~\cite{zheng2025diffusion, huang2023gameformer}. As a result, they struggle to generate trajectories that are simultaneously safe, adaptive, and aligned with user preferences. Moreover, users exhibit diverse safety perceptions and driving habits. For example, aggressive drivers may prefer tighter safety margins, while conservative users prioritize comfort and caution. Failing to adapt to these personalized preferences not only degrades ride comfort but also undermines user trust in autonomous systems. Therefore, trajectory planning must evolve from a purely safety-driven optimization problem into a user-centric decision task that jointly accounts for safety, comfort, and individualized driving styles~\cite{dauner2023pdm, zhang2024styledrive}.

Early efforts in personalized or stylized driving have progressed from handcrafted rule tuning to data-driven and preference-aware modeling methods~\cite{zhang2024styledrive, sima2024drivelm}. Handcrafted approaches use fixed parameters for simplicity but lack adaptability in dynamic contexts. Learning-based methods, especially reinforcement learning (RL) based approaches, can capture adaptive behavior patterns from data yet often sacrifice safety when balancing multiple objectives in a static manner~\cite{wu2024recent, dinneweth2022marlsurvey}. Preference-aware or benchmark-driven approaches~\cite{zhang2024styledrive} partially mitigate this issue by explicitly considering user style tendencies, but they still do not fully adapt in real time to evolving risks or low-risk conditions. Consequently, a gap remains for planners that dynamically reconcile safety and style under complex multi-agent environments while preserving interpretability and real-time feasibility.

\begin{figure}
    \centering
    \includegraphics[width=1\linewidth]{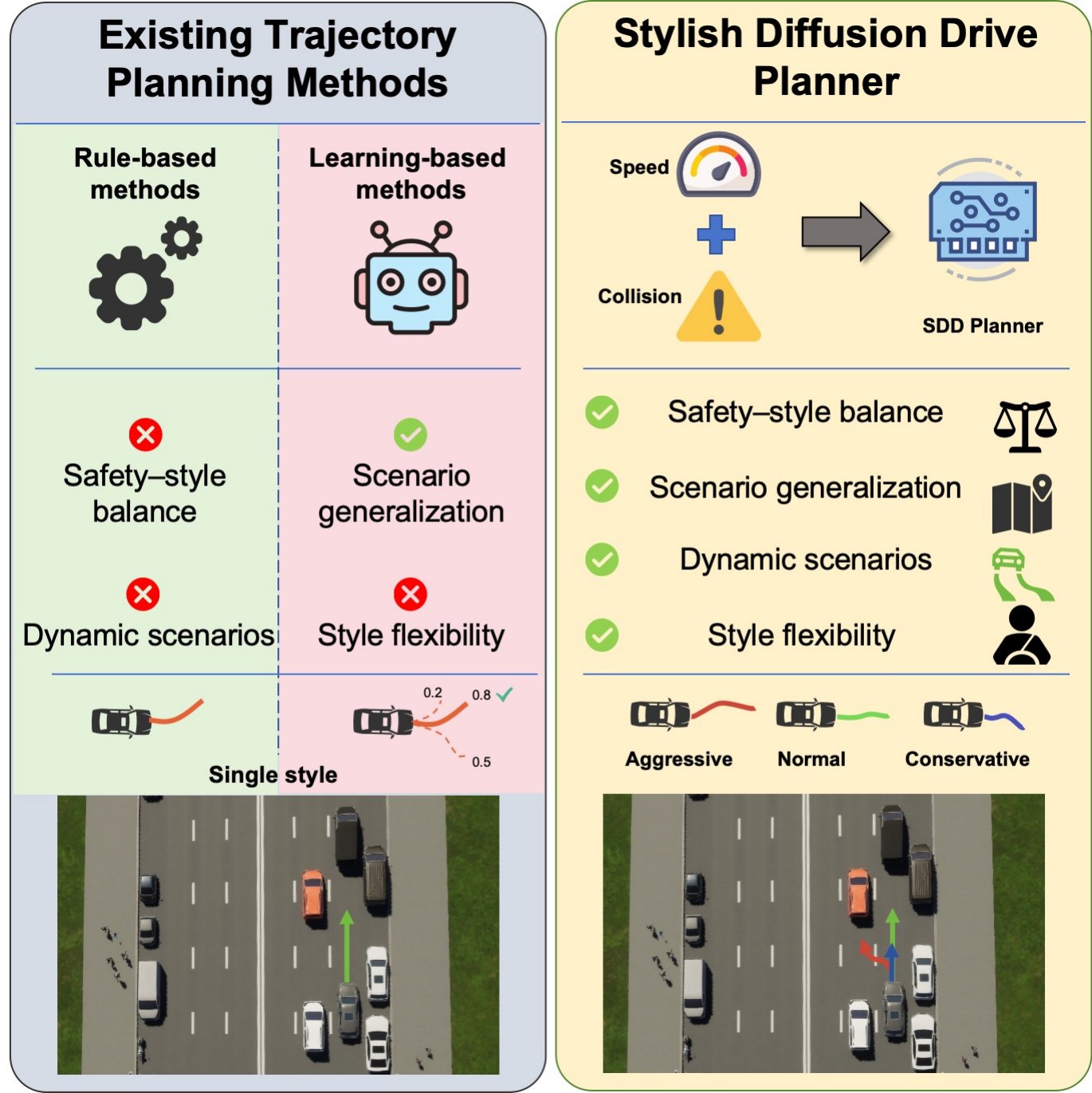}
    \caption{\textbf{Comparison of different trajectory planning paradigms}. Unlike existing methods that compromise between safety and style, the proposed SDD Planner achieves user-preferred, multi-style trajectories with adaptive safety performance, approaching the ideal balance point.}
    \label{fig:methods_comparison}
\end{figure}

To address these challenges, this paper proposes the \textbf{Stylish Diffusion Drive (SDD) Planner}, a diffusion-based trajectory planner designed for safe, adaptive, and stylized driving. As shown in Fig.~\ref{fig:methods_comparison}, existing methods often deviate from the ideal balance between safety adaptability and style flexibility, whereas the proposed SDD Planner generates trajectories closer to this ideal, flexibly adapting to multiple driving styles (aggressive, normal, conservative). The main contributions of this work are summarized as follows:

\begin{itemize}
    \item A Style-Driven Diffusion Planner (SDD Planner) is proposed for user-centric trajectory planning, achieving a dynamic balance between safety, comfort, and driving style through diffusion-based generative modeling.
    \item We design a Multi-Source Style-Aware Encoder with distance-sensitive attention, enabling precise fusion of dynamic agents and environmental context to accommodate heterogeneous safety and style preferences.
    \item We develop a Style-Guided Dynamic Trajectory Generator that adaptively adjusts priority weights in real time, producing trajectories that are both user-preferred and safe.
    \item Extensive experiments, including robustness analysis, attention/selection ablation, and closed-loop real-vehicle validation with a distilled deployment variant, verify that SDD Planner achieves strong safety compliance, style alignment, and preliminary real-world executability.
\end{itemize}

\section{RELATED WORK}
\label{sec:related_work}

\subsection{Rule-based Planning}
\label{subsec:rule-based_planning}
Rule-based planning methods represent a foundational paradigm for AD trajectory generation. These approaches rely on translating explicit traffic rules, handcrafted safety heuristics, and fixed operational boundaries into deterministic decision logic \cite{dauner2023pdm, caesar2023nuplan}. Their chief advantage is high stability and predictability in structured scenarios, where they ensure trajectory legality and basic collision avoidance by strictly adhering to predefined constraints, such as fixed safety buffers or speed limits \cite{li2025spatio}. However, the efficacy of rule-based methods is fundamentally constrained by their inherent rigidity. This rigidity creates a critical trade-off, making them ill-suited for balancing safety with stylized driving demands. First, their reliance on static, handcrafted parameters prevents adaptation to heterogeneous user preferences (e.g., aggressive vs. conservative profiles), as the fixed safety weights cannot be dynamically adjusted to individual driving styles \cite{dauner2023pdm, caesar2023nuplan}. Second, this same inflexibility leads to poor performance in complex, unstructured environments. The fixed rules often result in over-conservative behaviors that compromise comfort or fail to account for nuanced dynamic interactions with other agents \cite{huang2023gameformer}. Consequently, while rule-based methods provide a baseline for safety, they lack the adaptive flexibility required to co-optimize safety constraints with personalized, context-aware driving styles.

\subsection{Learning-based Planning}
\label{subsec:learning-based_planning}
Learning-based planning methods emerged to address the inflexibility of rule-based approaches, leveraging data-driven models to capture complex environmental interactions and stylistic driving patterns from large-scale datasets. This paradigm is broadly divided into modular and end-to-end subcategories, both aiming to enhance adaptability while preserving safety \cite{wu2024recent}. Modular learning methods decompose the planning problem into sequential sub-tasks (e.g., obstacle prediction, behavior classification, trajectory optimization), which are then optimized individually via supervised learning or reinforcement learning (RL). For instance, TransFuser \cite{chitta2022transfuser} fuses multi-modal perceptual features (LiDAR, camera) for end-to-end driving, while PlanTF \cite{cheng2023plantf} provides a strong imitation-based closed-loop planning baseline on nuPlan. While these methods improve adaptability, integrating style remains a challenge. Some learning-based approaches attempt to encode preference-related objectives or adaptive behavior through reinforcement learning and policy optimization \cite{wu2024recent, dinneweth2022marlsurvey}. However, they typically rely on static trade-offs between safety and preference objectives, failing to dynamically adjust priorities when risk levels change \cite{wu2024recent}. End-to-end learning methods directly map raw perceptual inputs to control commands. This approach eliminates handcrafted sub-modules, enabling the model to learn holistic, end-to-end driving styles. Pioneered by Bojarski et al. \cite{bojarski2016end} for steering control, this strategy was later extended by conditional imitation learning frameworks to generate style-aware trajectories based on expert demonstrations \cite{scheel2021urbandriver}. Despite their potential for style replication, end-to-end methods suffer from significant safety opacity. Their ``black-box'' nature makes it difficult to verify or guarantee safe behavior in out-of-distribution (OOD) scenarios. Consequently, style adaptation often inadvertently compromises safety, such as when replicating aggressive speed profiles that elevate collision risk \cite{dauner2023pdm, cheng2023plantf}. Overall, while learning-based methods show promise for stylized driving, they struggle to dynamically balance safety constraints with personalized preferences. Their main limitations, including fixed safety--style weights, weak dynamic agent modeling, and poor interpretability of end-to-end models, highlight the need for a unified approach that adapts safety in real time while maintaining flexible driving styles.

Recent studies in intelligent transportation have further highlighted the value of adaptive decision-making beyond conventional trajectory imitation. In particular, reinforcement learning based methods for connected vehicle control and lifelong energy management emphasize multi-scale optimization and transferability across varying operating conditions. These directions are complementary to our work: while they mainly focus on control or energy optimization, our method targets safe and stylized trajectory generation under coupled safety--style constraints.

Meanwhile, large language models (LLMs) have recently shown promise in autonomous driving-related reasoning, including high-level scene understanding, semantic task decomposition, and language-guided decision-making \cite{sima2024drivelm,shao2023lmdrive}. In contrast, the proposed SDD Planner focuses on low-level continuous trajectory generation with explicit safety and feasibility guidance. We believe LLM-based semantic priors and diffusion-based motion generation are complementary, and their integration is an interesting direction for future personalized driving systems.

\subsection{Diffusion-based Planning}
\label{subsec:diffusion_based_planning}
Diffusion-based generative models have recently emerged as a dominant paradigm in autonomous driving trajectory generation. Their primary advantage over deterministic learning-based methods lies in the capability to model multimodal trajectory distributions, thereby capturing diverse and physically plausible driving behaviors in dynamic environments \cite{sohl2015deep, chi2023diffusion, tang2024utdptp}. However, the application of these models to user-centric planning remains constrained by their rigidity in balancing safety and driving style \cite{janner2022diffuser, zheng2025diffusion}. Existing frameworks exhibit two primary limitations. First, most approaches lack dynamic reconciliation between safety constraints and stylistic preferences. Early works, such as Diffusion Policy \cite{chi2023diffusion}, focus on static goal-reaching tasks without explicit style awareness. While extensions like DiffusionES \cite{yang2024diffusiones} introduce multimodality via map conditioning, they typically rely on fixed safety weights, restricting adaptation to varying risk levels. Similarly, classifier-guided models \cite{zheng2025diffusion} often enforce safety constraints using relatively fixed parameters, failing to accommodate the heterogeneous safety margins required by aggressive versus cautious drivers \cite{zhang2024styledrive}. Second, environmental fusion and interaction modeling are often oversimplified. Diffusion-based planners may still under-model rapidly evolving multi-agent interactions when temporal dynamics and interaction structure are not sufficiently encoded \cite{zheng2025diffusion, huang2023gameformer}. Furthermore, some safety-guided or map-conditioned diffusion planners still rely on relatively simple guidance formulations and may not fully capture complex inter-agent influences in cluttered scenarios \cite{zheng2025diffusion, yang2024diffusiones}. These limitations highlight the need for a generative framework that can dynamically adapt to both safety requirements and user-aligned stylistic preferences \cite{zheng2025diffusion}.

\section{PRELIMINARIES}
\label{sec:preliminaries}

\subsection{Problem Formulation}
We formulate stylized trajectory planning as a conditional generative modeling task. Let $x^{(0)} \in \mathbb{R}^{B \times T_{\text{pred}} \times 2}$ denote the target ground-truth trajectory sequences, where $B$ represents the batch size and $T_{\text{pred}}$ denotes the prediction horizon. The last dimension corresponds to longitudinal and lateral coordinates in the ego-centric frame. The planning process is conditioned on a heterogeneous context set $\mathcal{I}$, which encompasses the ego-vehicle state, the positions of $N$ dynamic agents $\mathbf{p} \in \mathbb{R}^{B \times N \times 2}$, and static map elements. Furthermore, to explicitly govern the planning behavior, a discrete driving style indicator $S \in \{\textit{Aggressive}, \textit{Normal}, \textit{Conservative}\}$ is introduced. We embed the style indicator into a learnable style token $\mathbf{e}_S \in \mathbb{R}^{D}$ and jointly condition the planner on both scene context and style preference. Accordingly, the objective is to approximate the conditional distribution
\begin{equation}
p_{\theta}(x^{(0)} \mid \mathcal{I}, S),
\end{equation}
where the heterogeneous scene inputs and the style embedding are fused into a high-dimensional stylized representation $\boldsymbol{z}_{\text{style}}$.

\subsection{Conditional Diffusion Backbone}
To address the generative modeling of trajectory distributions, we adopt the Denoising Diffusion Probabilistic Model (DDPM) as the mathematical backbone. This framework approximates the data distribution $q(x^{(0)})$ via a parameterized Markov chain spanning $T$ diffusion steps \cite{sohl2015deep, ho2020denoising}.

The forward process gradually corrupts the clean trajectory $x^{(0)}$ by injecting Gaussian noise according to a fixed variance schedule $\beta_1, \dots, \beta_T$. At an arbitrary step $t$, the noisy state $x^{(t)}$ is sampled from the distribution:
\begin{equation}
    q(x^{(t)} | x^{(0)}) = \mathcal{N}(x^{(t)}; \sqrt{\bar{\alpha}_t}x^{(0)}, (1 - \bar{\alpha}_t)\mathbf{I}),
\end{equation}
where $\alpha_t = 1 - \beta_t$ and $\bar{\alpha}_t = \prod_{s=1}^t \alpha_s$ are coefficients derived from the variance schedule, and $\mathbf{I}$ denotes the identity matrix. As $t$ approaches the total steps $T$, the distribution of $x^{(T)}$ converges to a standard isotropic Gaussian $\mathcal{N}(0, \mathbf{I})$.

In the reverse process, a neural network $\epsilon_\theta$ is employed to reconstruct $x^{(0)}$ from the latent noise $x^{(T)}$ through iterative denoising. Crucially, this process is explicitly conditioned on the stylized embedding $\boldsymbol{z}_{\text{style}}$ to ensure the generated trajectories align with the driving context:
\begin{equation}
    p_\theta(x^{(t-1)} | x^{(t)}, \boldsymbol{z}_{\text{style}}) = \mathcal{N}(x^{(t-1)}; \mu_\theta(x^{(t)}, t, \boldsymbol{z}_{\text{style}}), \sigma_t^2 \mathbf{I}).
\end{equation}
Here, $\boldsymbol{z}_{\text{style}}$ represents the context features extracted by the encoder, the mean $\mu_\theta$ is predicted by the network $\epsilon_\theta(x^{(t)}, t, \boldsymbol{z}_{\text{style}})$, and $\sigma_t^2$ is a time-dependent variance term fixed to $\beta_t$ or $\tilde{\beta}_t$ \cite{ho2020denoising}. This formulation allows the high-dimensional features to guide the generation process foundationally.

\section{METHODOLOGY}
\label{sec:method}

\subsection{Overview}
\begin{figure*}[t]
    \centering
    \includegraphics[width=1\linewidth]{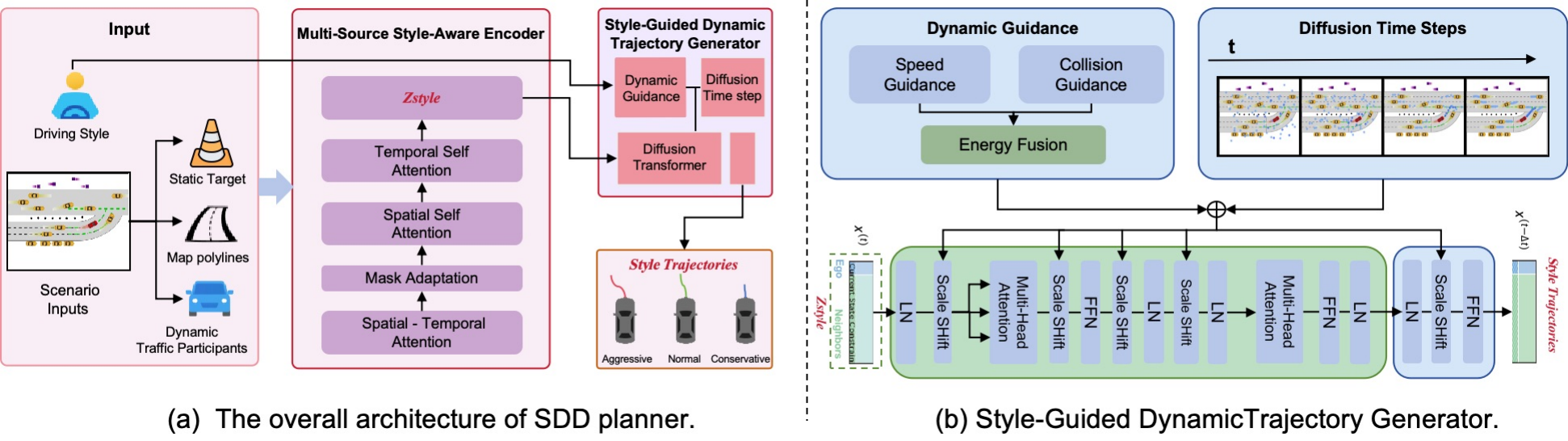}
    \caption{\textbf{Overall architecture of the SDD Planner.} (a) The framework encodes multi-source scenario inputs via a \textbf{Multi-Source Style-Aware Encoder} into high-dimensional stylized features ($\boldsymbol{z}_{style}$), serving as critical conditioning signals. (b) \textbf{Style-Guided Dynamic Trajectory Generator:} The diffusion decoder \textbf{iteratively denoises trajectories from Gaussian noise ($t=T$) to stylized paths ($t=0$)} via a time-step adaptive classifier-guided framework. This process is governed by: \textbf{Style Conditioning}, where $\boldsymbol{z}_{style}$ is injected via Scale Shift to enforce context consistency; and \textbf{Dynamic Guidance}, which modifies the score function via fused energy gradients to strictly enforce safety and style constraints.}
    \label{fig:fig2}
\end{figure*}

Fig.~\ref{fig:fig2} presents an overview of the proposed SDD Planner, which integrates multi-source style-aware encoding with style-guided dynamic trajectory generation. A diffusion model serves as the core of trajectory generation, enabling iterative denoising that aligns safety constraints and stylized preferences across time steps. To capture environmental information and user-specific style requirements, the Multi-Source Style-Aware Encoder processes the ego-vehicle state, dynamic traffic participants, static maps, traffic light states, and an explicit driving-style embedding $\mathbf{e}_S$. These inputs are transformed into unified feature representations using a distance-aware attention mechanism that prioritizes nearby participants according to stylized driving preferences. Temporal self-attention models cross-time dependencies, while spatial self-attention captures relative positional relationships; their fusion produces high-dimensional stylized features ($\boldsymbol{z}_{\text{style}}$) that encode both scene context and style attributes---serving as critical conditioning signals for the diffusion process. The core Style-Guided Dynamic Trajectory Generator consists of two components tightly coupled with the diffusion model: (1) A Dynamic Guidance Unit, which leverages classifier-guided diffusion with energy functions for collision avoidance and speed compliance, balancing safety and style expression during denoising; and (2) A Multi-Objective Fusion, Post-processing, and Decoding Unit, which refines, filters, and selects executable trajectories before final output. As shown in Fig.~\ref{fig:fig2}(b), the diffusion process progresses from $t=T$ (pure noisy trajectories) to $t=0$ (clean stylized trajectories), with each time step refining the trajectory to adhere to safety rules and match preset styles (aggressive, normal, conservative).

\subsection{Multi-Source Style-Aware Encoder}
In autonomous driving systems, the environmental perception encoder embeds multi-source heterogeneous inputs into a high-dimensional feature space. This space generates $\boldsymbol{z}_{\text{style}}$, which is fed into the diffusion decoder at every time step to ensure trajectory consistency with scene context and user preferences \cite{chitta2022transfuser}. In addition to scene observations, the encoder receives the discrete driving-style label $S$, which is mapped to a learnable embedding $\mathbf{e}_S$. This embedding is fused with the spatio-temporal scene features to form $\boldsymbol{z}_{\text{style}}$, so that the diffusion decoder is explicitly conditioned on both scene context and desired driving style. Key symbols and their physical meanings are defined in Table~\ref{tab:encoder_symbols}.

\begin{table}[htbp]
\centering
\caption{Key Symbols Definition for Multi-Source Style-Aware Encoder.}
\label{tab:encoder_symbols}
\renewcommand{\arraystretch}{1.2}
\resizebox{\linewidth}{!}{%
\scriptsize
\begin{tabular}{c c c c}
\hline
\textbf{Symbol} & \textbf{Physical Meaning} & \textbf{Dimension/Value} & \textbf{Data Source/Remark} \\
\hline
$B$ & Training batch size & $B=32$ & Model hyperparameter \\
$N$ & Dynamic traffic participants count & $N \leq 100$ & LiDAR/camera perception \\
$\mathbf{p}$ & Planar coordinates (x: fwd, y: lat) & $\mathbb{R}^{B \times N \times 2}$ & Ego-centric system \\
$T_{\text{pred}}$ & Prediction horizon (Time steps) & $T_{\text{pred}}=50$ ($5\,\text{s}$) & Planning parameter \\
$L_{\text{feat}}$ & Encoded feature sequence length & $L_{\text{feat}}=N \cdot T_{\text{pred}}$ & Spatio-temporal fusion \\
$D$ & Feature vector dimension & $D=128$ & Model hyperparameter \\
$H$ & Number of attention heads & $H=8$ & Transformer configuration \\
$\kappa$ & Distance attenuation coefficient & $\kappa \in (0,1]$ & Optimized on StyleDrive \\
$\gamma$ & Temporal attenuation coefficient & $\gamma=0.05$ & Cross-validation \\
\hline
\end{tabular}%
}
\end{table}

The Multi-Source Style-Aware Encoder introduces a distance-aware attention mechanism in relationship modeling, dynamically adjusting attention based on traffic participants' spatial positions to align with stylized requirements. This mechanism integrates a Euclidean distance-based bias in self-attention, prioritizing closer participants with higher weights.

For $N$ traffic participants with positions $\mathbf{p} \in \mathbb{R}^{B \times N \times 2}$, the pairwise Euclidean distance matrix $\mathbf{D}_{\text{dist}} \in \mathbb{R}^{B \times N \times N}$ is computed as:
\begin{equation}
\mathbf{D}_{\text{dist}}[b, i, j] = \| \mathbf{p}_{b,i} - \mathbf{p}_{b,j} \|_2,
\end{equation}
where $\| \cdot \|_2$ denotes the Euclidean norm ($L_2$ norm), and $\mathbf{D}_{\text{dist}}[b,i,j]$ represents the distance between the $i$-th and $j$-th participants in the $b$-th batch sample. For participants beyond the perception range, their distance is set to $\infty$ to avoid interfering with attention weight calculation.

A distance-aware attention bias is constructed using a learnable attenuation coefficient $\kappa > 0$:
\begin{equation}
\mathbf{M}_{\text{dist}} = -\kappa \cdot \mathbf{D}_{\text{dist}}.
\end{equation}
The negative sign ensures that distant participants have smaller bias values. When combined with softmax normalization, this reduces the attention weight of distant participants, ensuring $\boldsymbol{z}_{\text{style}}$ emphasizes critical dynamic agents (e.g., nearby vehicles) that directly influence diffusion-based trajectory safety. The coefficient $\kappa$ is initialized to $0.1$ and constrained to $(0,1]$ during training. Empirically, $\kappa$ exhibits a stable operating range in cross-dataset evaluation. Our robustness analysis on NuPlan shows that directly transferring the StyleDrive-optimized $\kappa$ remains functional but introduces moderate degradation, whereas a mid-range setting near the default value provides the best trade-off among safety, efficiency, and smoothness. This suggests that the proposed distance-sensitive attention is robust within a reasonable range of $\kappa$, while lightweight scenario- or domain-adaptive calibration could further improve generalization across datasets.

Extended to multi-head attention (with $H$ attention heads), the bias is replicated along the head dimension to form the final attention mask:
\begin{equation}
\mathbf{M}_{\text{head}} = \mathbf{M}_{\text{dist}} \otimes \mathbf{1}_{H},
\end{equation}
where $\otimes$ denotes replication along the head dimension, and $\mathbf{M}_{\text{head}} \in \mathbb{R}^{B \times H \times N \times N}$ adjusts the attention weight of each head to ensure consistent distance-aware guidance.

Invalid participants (e.g., out-of-range objects, perception confidence $<0.6$) are masked to $-10^9$ via a binary mask $\mathbf{mask} \in \{0,1\}^{B \times N}$. To align with the attention dimensions, this mask is broadcasted to $\mathbb{R}^{B \times H \times N \times N}$:
\begin{equation}
\mathbf{M}_{\text{final}} = \mathbf{M}_{\text{head}} + \text{Broadcast}(\mathbf{mask}) \cdot (-10^9),
\end{equation}
where invalid interactions in rows and columns are penalized to eliminate interference.

The temporal self-attention matrix $\mathcal{A}_t \in \mathbb{R}^{T_{\text{pred}} \times T_{\text{pred}}}$ models cross-time behavioral dependencies, integrating spatial distances and temporal weights to capture how dynamic agents' movements evolve over time:
\begin{equation}
\mathcal{A}_t[i,j] = \exp\left(-\gamma \| \mathbf{v}_i - \mathbf{v}_j \|^2 + \mathbf{r}_{\text{temporal}}[i,j] \right),
\end{equation}
where $\mathbf{v}_i$ denotes the feature vector of the $i$-th time step (fused ego-vehicle state and dynamic agent features with dimension $D=128$); $\gamma=0.05$ is a temporal attenuation coefficient balancing short- and long-term time dependencies; and $\mathbf{r}_{\text{temporal}}[i,j]$ is a learnable temporal relative positional embedding adopting sinusoidal encoding. Sensitivity analysis further indicates that $\gamma=0.05$ provides the most balanced behavior across safety, progress, and comfort metrics. Smaller values weaken temporal discrimination and tend to produce overly conservative or less stable interactions, whereas larger values over-emphasize local temporal changes and reduce trajectory smoothness.

In the spatial dimension, the self-attention matrix $\mathcal{A}_s \in \mathbb{R}^{N \times N}$ allocates weights using relative position features to encode spatial relationships between the ego-vehicle and obstacles:
\begin{equation}
\mathcal{A}_s[m,n] = \sigma \left( \mathbf{W}_s \left[ \mathbf{v}_m \| \mathbf{v}_n \| \mathbf{r}_{\text{emb}} \right] \right),
\end{equation}
where $\sigma$ is the sigmoid function, and $\mathbf{W}_s \in \mathbb{R}^{D \times 3D}$ is a learnable weight matrix. Here, $\mathbf{r}_{\text{emb}} \in \mathbb{R}^D$ is the embedding of the relative position $\mathbf{r}_{a2a} = (\mathbf{p}[m] - \mathbf{p}[n])/100$, projected via an MLP to match the feature dimension $D$.

Temporal and spatial attention matrices are fused via Kronecker product to integrate multi-dimensional spatiotemporal relationships:
\begin{equation}
\mathcal{A}_{\text{fusion}} = \mathcal{A}_t \otimes \mathcal{A}_s \in \mathbb{R}^{(N \cdot T_{\text{pred}}) \times (N \cdot T_{\text{pred}})}.
\end{equation}
The Kronecker product ensures that the fused matrix retains both temporal dependencies and spatial relationships. The fused attention matrix refines multi-source input features into $\boldsymbol{z}_{\text{style}} \in \mathbb{R}^{B \times L_{\text{feat}} \times D}$, ensuring the diffusion decoder receives scene-aware and style-aligned guidance.

\subsection{Style-Guided Dynamic Trajectory Generator}
To enable stylized autonomous driving with safety constraints, we design a time-step adaptive classifier-guided diffusion framework. Its core modifies the diffusion model's score function via energy gradients, with guidance strength and energy weights adjusted dynamically across diffusion time steps.

\subsubsection{Diffusion Model Foundation: Forward and Reverse Processes}
The diffusion model operates over $T$ discrete time steps (set to $T=1000$), divided into two phases. The diffusion backbone is trained with the standard $T=1000$-step formulation to ensure stable learning of multimodal trajectory distributions. For real-vehicle inference, however, we employ DDIM-based fast sampling with only $N=15$ denoising steps, which substantially reduces runtime while preserving trajectory quality and diversity.

Forward Diffusion (Noising Process): Starting from a clean trajectory $\boldsymbol{x}^{(0)} \in \mathbb{R}^{B \times L \times 2}$ (where the last dimension represents longitudinal and lateral coordinates), we iteratively add Gaussian noise to generate a noisy trajectory $\boldsymbol{x}^{(t)}$ at time step $t$:
\begin{equation}
\boldsymbol{x}^{(t)} = \sqrt{\bar{\alpha}_t} \boldsymbol{x}^{(0)} + \sqrt{1 - \bar{\alpha}_t} \boldsymbol{\epsilon},
\end{equation}
where $\boldsymbol{\epsilon} \sim \mathcal{N}(0, \mathbf{I})$ is standard Gaussian noise, $\bar{\alpha}_t = \prod_{s=1}^t \alpha_s$ denotes the cumulative product of noise coefficients, and $\alpha_s$ follows a pre-defined linear variance schedule. This schedule ensures that by $t=T$, $\boldsymbol{x}^{(T)}$ approximates pure Gaussian noise.

Reverse Diffusion (Denoising Process): The diffusion decoder learns to reverse the noising process, predicting the noise $\boldsymbol{\epsilon}_\theta(\boldsymbol{x}^{(t)}, t, \boldsymbol{z}_{\text{style}})$ at each time step $t$, conditioned on the noisy trajectory $\boldsymbol{x}^{(t)}$, time step $t$, and stylized features $\boldsymbol{z}_{\text{style}}$. The trajectory is updated as:
\begin{equation}
\boldsymbol{x}^{(t-1)} = \frac{1}{\sqrt{\alpha_t}} \left( \boldsymbol{x}^{(t)} - \frac{1 - \alpha_t}{\sqrt{1 - \bar{\alpha}_t}} \boldsymbol{\epsilon}_\theta \right) + \sigma_t \boldsymbol{\epsilon}',
\end{equation}
where $\sigma_t$ is the time-step-dependent noise scale derived via Bayesian theorem ($\sigma_t^2 = \frac{1 - \bar{\alpha}_{t-1}}{1 - \bar{\alpha}_t} \cdot (1 - \alpha_t)$), and $\boldsymbol{\epsilon}' \sim \mathcal{N}(0, \mathbf{I})$ is residual noise to preserve trajectory multimodality. As $t$ decreases from $T$ to $0$, $\boldsymbol{x}^{(t)}$ gradually converges to a clean, stylized trajectory.

\subsubsection{Training Objective and Inference Usage}
The diffusion backbone is trained with the standard noise-prediction objective:
\begin{equation}
\mathcal{L}_{\text{diff}} =
\mathbb{E}_{x^{(0)},\,\epsilon,\,t}
\left[
\left\|
\epsilon - \epsilon_\theta\!\left(x^{(t)}, t, \boldsymbol{z}_{\text{style}}\right)
\right\|_2^2
\right].
\end{equation}
Here, $x^{(t)}$ is obtained from the forward diffusion process and $\boldsymbol{z}_{\text{style}}$ provides the scene-style conditioning. During training, the model learns the conditional denoising backbone through $\mathcal{L}_{\text{diff}}$. During inference, safety- and speed-related energies are additionally incorporated through guidance gradients to steer the reverse denoising process toward safe and style-consistent trajectories. Unless otherwise stated, the benchmark experiments use the same trained diffusion backbone, while DDIM-based reduced-step sampling is specifically adopted for real-vehicle deployment to meet runtime requirements.

\subsubsection{Dynamic Guidance Across Diffusion Time Steps}
To integrate safety and style into the denoising process, we modify the denoiser's score function with time-step adaptive energy guidance:
\begin{equation}
\tilde{s}_\theta(\boldsymbol{x}^{(t)},t) = \boldsymbol{s}_\theta(\boldsymbol{x}^{(t)},t) - \lambda(t)\nabla_{\boldsymbol{x}^{(t)}}\mathcal{E}_\phi(\boldsymbol{x}^{(t)},t),
\end{equation}
where $\boldsymbol{s}_\theta(\boldsymbol{x}^{(t)},t) = -\nabla_{\boldsymbol{x}^{(t)}} \log p_\theta(\boldsymbol{x}^{(t)})$ denotes the base score function of the diffusion model, and $\mathcal{E}_\phi(\boldsymbol{x}^{(t)},t)$ is the fused guidance energy. To prioritize safety in the early reverse denoising stage and gradually relax the guidance later, we define
\begin{equation}
\lambda(t) = 0.3 + 1.2 \cdot \frac{t}{T},
\end{equation}
so that stronger guidance is applied when $t$ is large (early denoising) and weaker guidance is applied when $t$ approaches $0$ (late denoising). This schedule is motivated by the staged behavior of diffusion denoising: early steps mainly shape coarse global feasibility, while later steps refine fine-grained style and comfort attributes. This property is particularly desirable for autonomous driving, where collision-free safety should be established before style-specific refinements are introduced.

\subsubsection{Dynamic-Weight Collision Avoidance Guidance}
For safety-style alignment, a collision risk energy function quantifies the probability of collision for $\boldsymbol{x}^{(t)}$ at each time step:
\begin{equation}
\mathcal{E}_{\text{collision}}(\boldsymbol{x}^{(t)}) = \sum_{k=0}^{L-1} \sum_i \exp\left(-\frac{d_i^2(t,k)}{\sigma_d^2}\right),
\end{equation}
where $d_i(t,k)$ denotes the minimum Euclidean distance between the ego-vehicle and the $i$-th obstacle at trajectory point $k$; $\sigma_d = 2.5\,\text{m}$ is the distance attenuation coefficient determined via statistical analysis; and $L$ represents the number of trajectory points. The exponential term ensures that closer obstacles lead to higher energy penalties.

A multi-risk fusion weight function adjusts the collision avoidance priority across diffusion time steps:
\begin{equation}
\begin{split}
w_{\text{collision}}(t) &= \alpha(t) \cdot \sum_i \frac{1}{d_i(t) + \epsilon} \cdot \max\left(0, \frac{v_{\text{rel},i}(t)}{v_{\text{max,rel}}}\right) \\
&\quad \cdot \exp\left(\frac{c(t)}{\sigma_c}\right),
\end{split}
\end{equation}
where $\alpha(t) = \alpha_0 \cdot (1 + 0.8 \cdot \frac{t}{T})$ is the safety base weight; $v_{\text{rel},i}(t)$ denotes the relative speed of the $i$-th obstacle; and $c(t)$ represents the road curvature at time step $t$. This function enables adaptive response to high-risk scenarios.

\subsubsection{Dynamic-Weight Speed Compliance Guidance}
For style-specific speed control, a speed energy function measures deviation from style benchmarks at each diffusion time step:
\begin{equation}
\mathcal{E}_{\text{speed}}(\boldsymbol{x}^{(t)}) = \sum_{k=0}^{L-1} \left( \frac{v_{\text{current}}(t,k) - v_{\text{desired}}(t,k)}{v_{\text{limit}}(t)} \right)^2,
\end{equation}
where $v_{\text{current}}(t,k)$ is the speed of the ego-vehicle; $v_{\text{desired}}(t,k)$ denotes the style-specific benchmark speed (e.g., $1.1 \cdot v_{\text{limit}}$ for aggressive style); and $v_{\text{limit}}(t)$ represents the road speed limit.

A style-adaptive speed weight function adjusts priority across time steps:
\begin{equation}
\begin{split}
w_{\text{speed}}(t) &= \beta(t) \cdot \min\left(1, \frac{|\Delta v(t)|}{v_{\text{desired}}(t)}\right) \\
&\quad \cdot \left(1 + \gamma \cdot \exp\left(-\frac{\rho(t)}{\sigma_\rho}\right)\right),
\end{split}
\end{equation}
where $\beta(t) = \beta_0 \cdot (1 - 0.6 \cdot \frac{t}{T})$ is the speed base weight; $\Delta v(t)$ denotes the speed deviation; and $\rho(t)$ represents the traffic density at time step $t$.

\subsubsection{Multi-Objective Energy Fusion}
A normalized energy function fuses collision avoidance and speed compliance signals, ensuring stable guidance across diffusion time steps:
\begin{equation}
\begin{split}
\mathcal{E}(\boldsymbol{x}^{(t)}) &= \frac{w_{\text{collision}}(t)}{w_{\text{collision}}(t) + w_{\text{speed}}(t)} \mathcal{E}_{\text{collision}}(\boldsymbol{x}^{(t)}) \\
&\quad + \frac{w_{\text{speed}}(t)}{w_{\text{collision}}(t) + w_{\text{speed}}(t)} \mathcal{E}_{\text{speed}}(\boldsymbol{x}^{(t)}).
\end{split}
\end{equation}
This fusion enables scenario-adaptive priority adjustment, ensuring that the Style-Guided Dynamic Trajectory Generator dynamically reconciles safety constraints and user preferences throughout the denoising process.

\subsubsection{Trajectory Post-processing, Scoring and Selection}
Given a set of $K$ candidate trajectories generated by the diffusion model, we first apply a feasibility-oriented post-processing module to correct or remove trajectories that violate kinematic or safety-related constraints. Among the remaining candidates, the final output is selected by minimizing a composite score:
\begin{equation}
\mathcal{J}_{\text{sel}}(\tau_i)
=
\eta_{\text{col}}\,\mathcal{C}_{\text{col}}(\tau_i)
+
\eta_{\text{spd}}\,\mathcal{C}_{\text{spd}}(\tau_i)
+
\eta_{\text{comf}}\,\mathcal{C}_{\text{comf}}(\tau_i),
\end{equation}
where $\mathcal{C}_{\text{col}}$, $\mathcal{C}_{\text{spd}}$, and $\mathcal{C}_{\text{comf}}$ denote collision-related, speed-related, and comfort-related costs, respectively. The final trajectory is selected as
\begin{equation}
\tau^\ast = \arg\min_{\tau_i \in \mathcal{T}_{\text{valid}}} \mathcal{J}_{\text{sel}}(\tau_i),
\end{equation}
where $\mathcal{T}_{\text{valid}}$ denotes the candidate set after feasibility-oriented post-processing. This strategy combines multimodal diffusion generation with practical constraint handling for closed-loop execution. Under extreme risk, safety dominates the optimization priority, while style is preserved within the valid candidate set through the persistent style conditioning and the late-stage denoising refinement.

\section{EXPERIMENT}
\label{sec:experiment}

\subsection{Experimental Setup}
\subsubsection{Datasets}
Experiments are conducted on two authoritative autonomous driving datasets: StyleDrive \cite{zhang2024styledrive} and NuPlan \cite{caesar2023nuplan}. \textbf{StyleDrive} is a real-world dataset for personalized end-to-end autonomous driving (E2EAD), designed to benchmark driving-style-aware planning and evaluation. It provides multi-source style labels (e.g., aggressive, normal, conservative) together with scene topology, semantics, and trajectory data, serving as a benchmark for personalized E2EAD models. \textbf{NuPlan} provides large-scale real-world driving data with multi-modal sensor streams, HD maps, and challenging interaction events. It includes standard benchmark splits (Val14, Test14, and Test14-hard) for model validation and evaluation in regular and high-risk scenarios.

\subsubsection{Evaluation Metrics}
To assess whether the proposed model achieves a balance between safety/feasibility and alignment with driving style preferences, we adopt the \textbf{Style-Modulated Predictive Driver Model Score (SM-PDMS)} \cite{zhang2024styledrive}. SM-PDMS extends the Predictive Driver Model Score (PDMS) by introducing a behavior alignment term that measures the consistency between the policy and specified style preferences. The metric retains key PDMS components, including \textbf{No Collision (NC)} and \textbf{Drivable Area Compliance (DAC)}. We further adopt a set of standard closed-loop indicators, including \textbf{Collisions}, \textbf{Time to Collision (TTC)}, \textbf{Drivable}, \textbf{Comfort}, \textbf{Progress (EP)}, and the aggregated \textbf{Score}, which together quantify safety, rule compliance, ride comfort, and driving efficiency. All performance results are reported as normalized scores (0--100), averaged across all scenarios.

\subsubsection{SOTA Methods}
We compare the proposed SDD Planner with representative state-of-the-art baselines:
\textbf{AD-MLP} \cite{zhai2023rethinking}: A base end-to-end autonomous driving (E2EAD) model with an MLP core, lacking style adaptation.
\textbf{TransFuser} \cite{chitta2022transfuser}: A multi-modal E2EAD model fusing image and LiDAR features, generating only generic trajectories.
\textbf{WoTE} \cite{li2025wote}: A BEV-based E2EAD baseline focusing on long-term environment prediction, lacking explicit style adaptation.
\textbf{Diffusion-Planner} \cite{zheng2025diffusion}: A trajectory planning model based on diffusion probabilistic models.
\textbf{IDM} \cite{treiber2000intelligent}: A classical rule-based planner officially implemented on the NuPlan platform.
\textbf{PDM} \cite{dauner2023pdm}: A strong nuPlan planning baseline combining rule-based priors and learning-based strategies.
\textbf{UrbanDriver} \cite{scheel2021urbandriver}: A conditional-imitation-learning-based planner for urban driving scenarios.
\textbf{GameFormer} \cite{huang2023gameformer}: A game-theoretic interaction model refined with rule-based conflict resolution.
\textbf{PlanTF} \cite{cheng2023plantf}: An imitation-based planner designed for strong closed-loop planning performance.
\textbf{PLUTO} \cite{cheng2024pluto}: A state-of-the-art imitation learning-based planner that utilizes closed-loop reasoning to push the limits of autonomous driving planning.

\subsubsection{Implementation Details}
All experiments were conducted on a high-performance server. The details of the hardware and software configurations used in the experiment are as follows: In terms of hardware, the Central Processing Unit (CPU) is Intel (R) Xeon (R) Platinum 8362, with an operating frequency of 2.80 GHz; the Graphics Processing Unit (GPU) adopts the NVIDIA A100-SXM4 model and is equipped with 4 sets of 80 GB GPU memory. For software, the operating system uses Ubuntu 22.04, the CUDA Toolkit version is 12.1, and the deep learning framework adopts PyTorch 2.0.0+cu118.

\subsection{Quantitative Results}

\begin{table}[h]
\centering
\caption{Performance comparison on the StyleDrive dataset.
SDD-A, SDD-N, and SDD-C denote results on Aggressive, Normal, and Conservative driving styles, respectively.}
\label{tab:styledrive_performance}
\renewcommand{\arraystretch}{1.1}
\scriptsize
\begin{tabular}{lcccccc}
\hline
{Models} & {NC} & {DAC} & \multicolumn{3}{c}{Style-Modulated Submetrics} & {SM-PDMS} \\
\cline{4-6}
& & & TTC & Comf. & EP \\
\hline
AD-MLP  & 92.44 & 78.35 & 83.67 & 99.62 & 77.91 & 63.70 \\
TransFuser   & 96.54 & 87.92 & 90.96 & 99.70 & 84.46 & 78.13 \\
WoTE  & 97.32 & 92.29 & 92.60 & 99.05 & 76.52 & 80.32 \\
Diffusion-Planner & 96.50 & 91.65 & 90.67 & 99.72 & 80.34 & 80.21 \\
\hline
\rowcolor{gray!30}
SDD (ours) & \textbf{97.81} & \textbf{93.50} & \textbf{92.79} & \textbf{99.87} & \textbf{84.79} & \textbf{84.23} \\
\rowcolor{gray!10}
SDD-A (ours) & 97.34 & 92.95 & 91.86 & 99.50 & 84.11 & 82.92 \\
\rowcolor{gray!10}
SDD-N (ours) & 97.65 & 93.30 & 92.11 & 99.71 & 84.19 & 83.32 \\
\rowcolor{gray!10}
SDD-C (ours) & 98.26 & 93.45 & 94.99 & 99.86 & 81.53 & 83.91 \\
\hline
\end{tabular}
\end{table}

\begin{table*}[h]
\renewcommand{\arraystretch}{1.05}
\caption{Performance comparison on NuPlan dataset.}
\label{tab:nuplan_combined}
\centering
\scriptsize
\begin{tabular}{lcc|cccccc}
\hline
\textbf{Planner (Type)} & \textbf{Test14-hard} & \textbf{Test14} & \multicolumn{6}{c}{\textbf{Val14}} \\
\cline{2-3} \cline{4-9}
& \textbf{Score} & \textbf{Score} & \textbf{Score} & \textbf{Collisions} & \textbf{TTC} & \textbf{Drivable} & \textbf{Comfort} & \textbf{Progress} \\
\hline
IDM        & 56.21 & 70.42 & 75.43 & 86.12 & 78.93 & 99.41 & 89.00 & 95.43 \\
PDM-Closed & 65.08 & 91.03 & 91.72 & 86.51 & 87.10 & 100.0 & 97.21 & 93.47 \\
PDM-Hybrid & 66.13 & 90.28 & 90.71 & 86.50 & 76.20 & 100.0 & 92.51 & 98.12 \\
GameFormer & 67.85 & 81.85 & 79.78 & 82.50 & 72.50 & 65.00 & 98.00 & 90.00 \\
PLUTO      & 80.12 & 91.29 & 90.68 & 89.95 & 87.64 & 99.45 & 94.47 & 97.79 \\
PDM-Open*    & 35.53 & 52.23 & 54.24 & 65.75 & 69.50 & 93.50 & 98.50 & 95.00 \\
UrbanDriver  & 49.95 & 57.15 & 54.11 & 67.32 & 70.84 & 95.34 & 92.15 & 94.76 \\
GameFormer w/o refine.  & 7.08 & 11.31 & 12.69 & 42.20 & 27.50 & 23.00 & \textbf{98.53} & 57.00 \\
PlanTF      & 69.63 & 85.58 & 84.75 & 89.00 & 87.50 & 99.50 & 96.00 & 99.50 \\
PLUTO w/o refine.*  & 70.74 & 89.62 & 88.11 & 92.39 & 87.61 & 99.54 & 97.19 & 98.78 \\
\rowcolor{gray!20}
SDD Planner (ours) & \textbf{80.32} & \textbf{91.76} & \textbf{91.83} & \textbf{96.00} & \textbf{90.79} & 100.0 & 94.00 & \textbf{100.0} \\
\hline
\end{tabular}
\end{table*}

As summarized in Table~\ref{tab:styledrive_performance}, the proposed SDD Planner consistently surpasses baseline models (AD-MLP, TransFuser, WoTE, Diffusion-Planner) across both safety and style-modulated evaluation metrics on the StyleDrive dataset. In terms of safety, SDD Planner achieves the highest scores in NC and DAC, improving by 0.51\% and 1.22\% compared to the strongest baseline. For style-sensitive metrics, TTC remains competitive, while SDD Planner attains leading performance in both Comf. (99.87) and EP (84.79), highlighting its strength in balancing efficiency and comfort. Most notably, the integrated SM-PDMS score demonstrates a substantial 3.9\% gain over the strongest baseline, confirming SDD Planner’s superior overall capability in stylized autonomous driving. When evaluated under fixed driving style constraints, the SDD variants reveal clear alignment with intended behavioral tendencies. Specifically, SDD-A shows minimal EP degradation while maintaining efficiency priority; SDD-N yields balanced yet slightly reduced results across metrics; and SDD-C improves safety-critical metrics at the cost of lower efficiency. These shifts precisely reflect aggressive, normal, and conservative driving preferences, thereby validating the adaptability of the proposed planner to diverse styles without compromising core safety requirements. Overall, these findings demonstrate that SDD Planner achieves a dual objective: ensuring robust safety guarantees while flexibly adapting to personalized driving styles. This balance between safety feasibility and preference alignment underscores its practicality and scalability for real-world deployment in complex, style-diverse autonomous driving scenarios.

As shown in Table~\ref{tab:nuplan_combined}, the proposed SDD Planner achieves superior performance across the NuPlan benchmarks, consistently outperforming state-of-the-art hybrid, learning-based, and rule-based methods in both overall score and critical sub-indicators. In the \textbf{Test14-hard} benchmark, which evaluates planners under highly challenging driving conditions, SDD Planner achieves the best overall score (80.32), slightly surpassing PLUTO (80.12). In the \textbf{Test14} benchmark, SDD Planner again ranks first with a score of 91.76, outperforming PLUTO (91.29) and all other methods, confirming its stability in standard urban driving conditions. In the \textbf{Val14} benchmark, SDD Planner ranks first with an overall score of \underline{91.83}. Compared with PLUTO, the leading imitation-learning-based planning method (90.68), SDD Planner achieves substantial gains in safety-related indicators, reducing collisions by 6.05\% and improving TTC by 3.15\%, while also obtaining a perfect score of 100.0 in the progress metric. Against PDM-Closed, the strongest hybrid baseline (91.72), SDD Planner achieves a slightly higher overall score but significantly outperforms it in safety-critical dimensions, including collisions and TTC. These results confirm that SDD Planner achieves state-of-the-art performance on the NuPlan benchmarks. It not only achieves strong efficiency but also ensures superior safety guarantees.

\subsection{Case Analysis}

\begin{figure}[ht]
    \centering
    \includegraphics[width=1\linewidth]{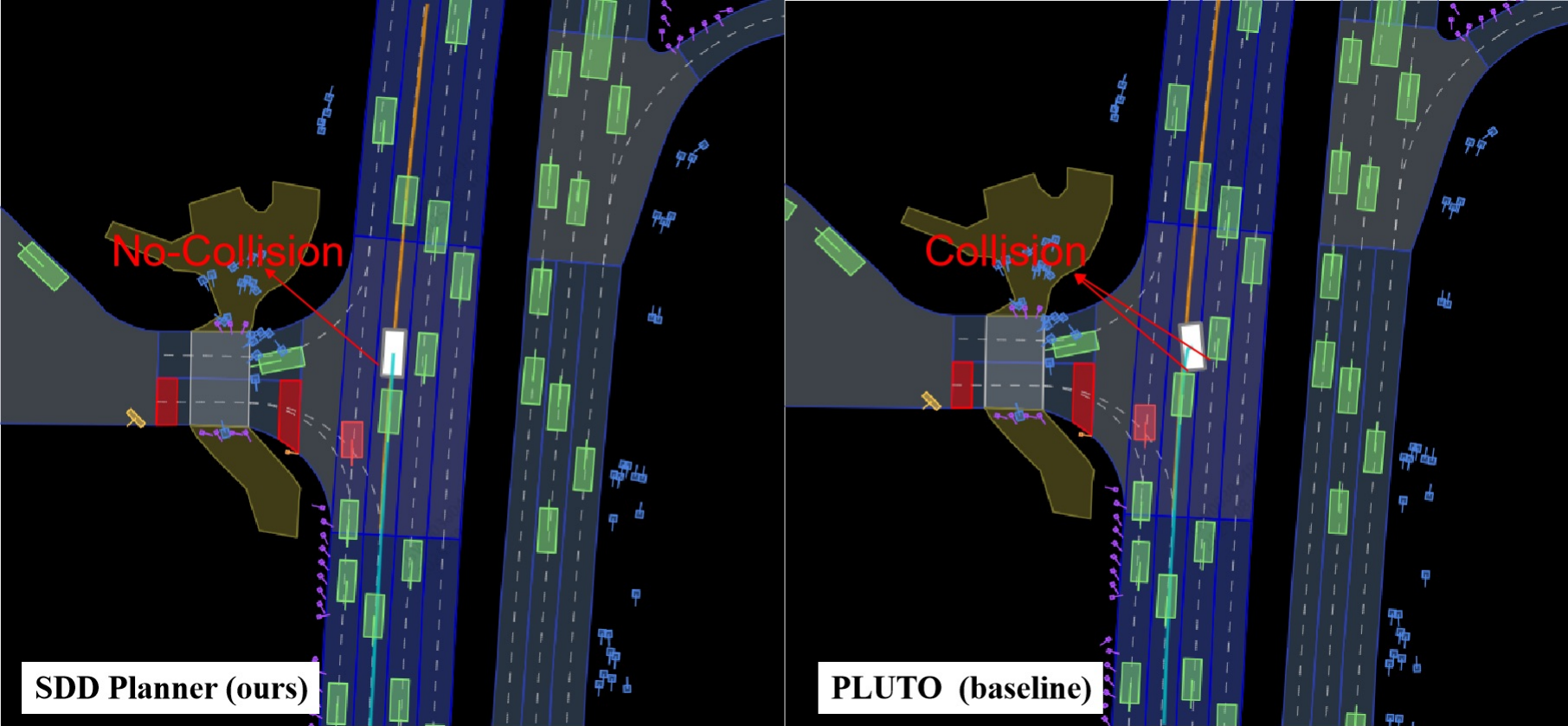}
    \caption{The performance of SDD Planner and PLUTO in the same scenario.}
    \label{fig:Qualitative_evaluation}
\end{figure}

\begin{figure}
    \centering
    \includegraphics[width=1\linewidth]{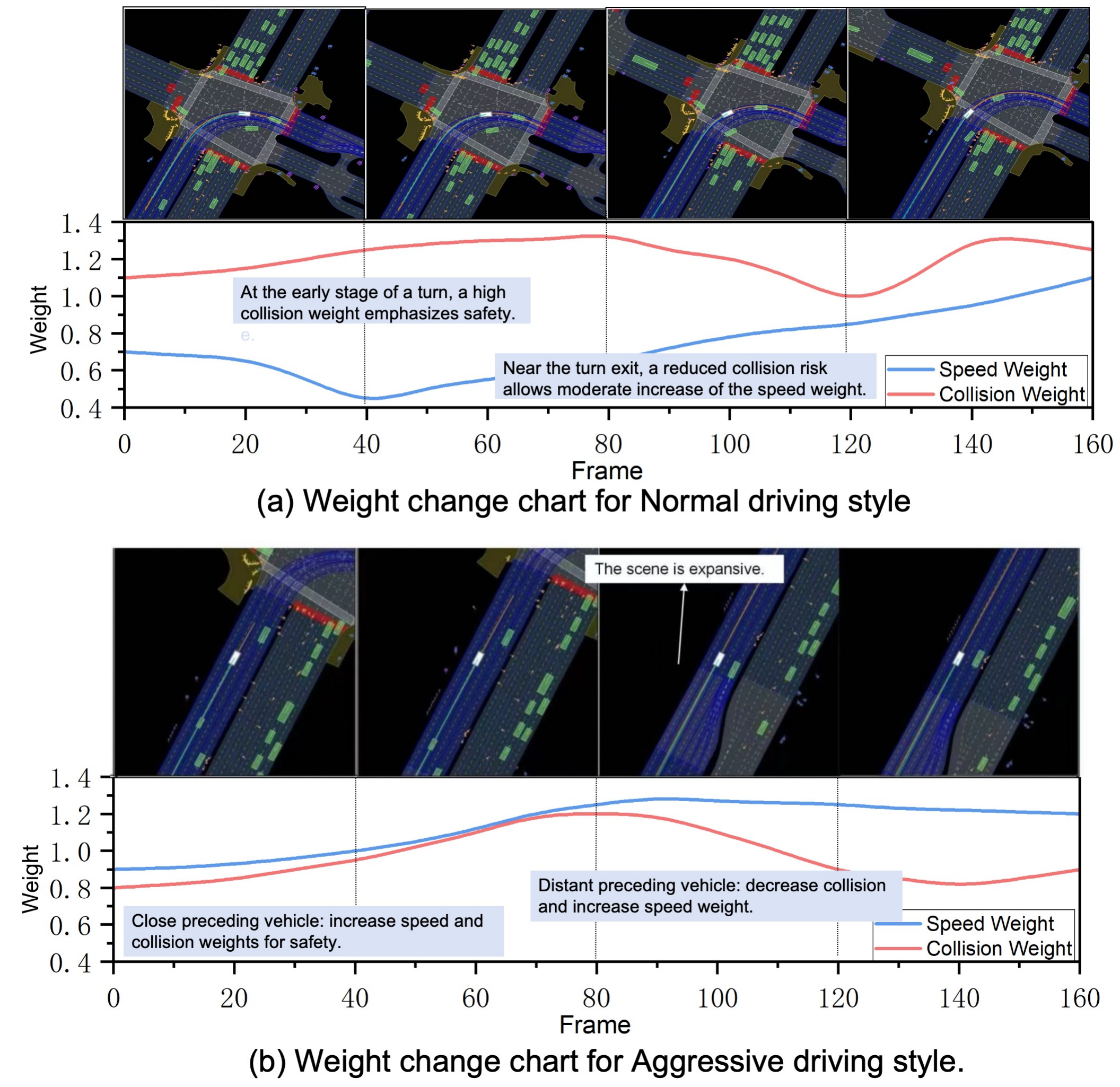}
    \caption{Weight change chart for two different driving scenarios.}
    \label{fig:diff_weight}
\end{figure}

\begin{figure}[ht]
    \centering
    \includegraphics[width=1\linewidth]{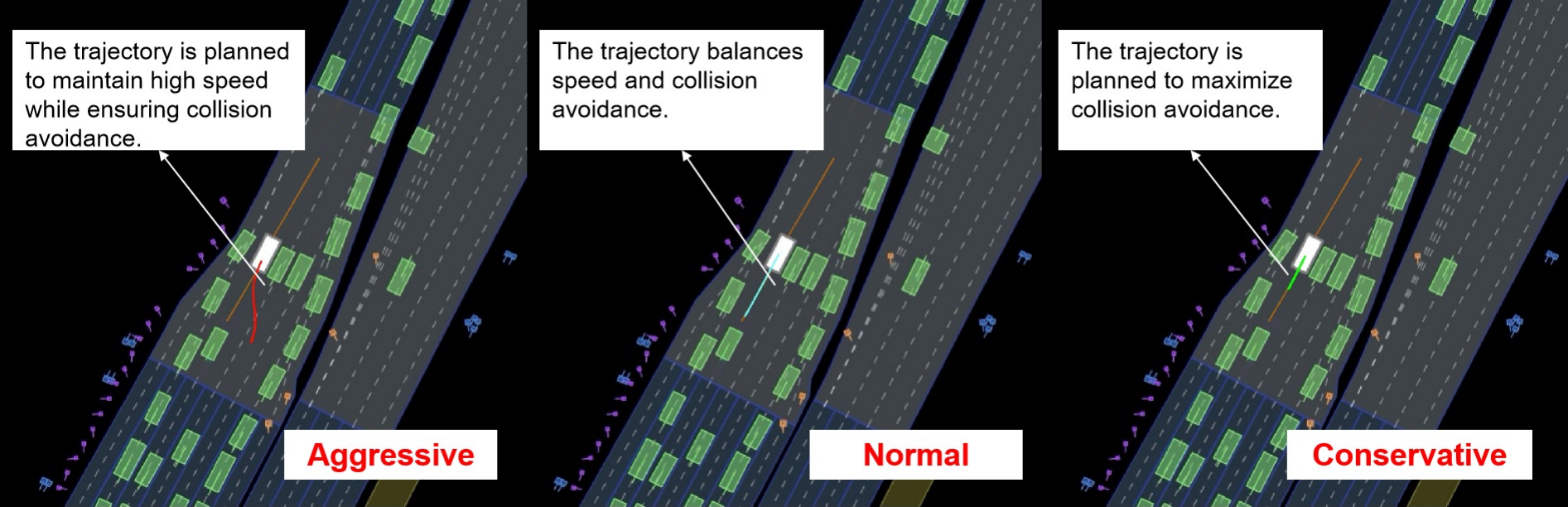}
    \caption{Comparison of generated trajectories under different driving styles.}
    \label{fig:ANC_trajectory}
\end{figure}

This subsection provides a visualization analysis to further verify the safety and personalization capabilities of the proposed SDD Planner. As shown in Fig.~\ref{fig:Qualitative_evaluation}, in a representative scenario from the Test14-hard benchmark, PLUTO fails to maintain safe interactions with surrounding vehicles, as it repeatedly shifts attention between the preceding and adjacent vehicles. This strategy leads to delayed decision-making and eventually a collision. In contrast, SDD Planner leverages its encoder to focus on the most critical target vehicle, enabling timely and stable interaction, thereby ensuring both safety and comfort.

Fig.~\ref{fig:diff_weight} visualizes the adaptive adjustment of dynamic weights under different driving styles. In the normal style case (Fig.~\ref{fig:diff_weight}(a)), as the driving scene becomes more expansive, the speed weight increases while the collision weight decreases, highlighting a preference for efficiency. In the aggressive style case (Fig.~\ref{fig:diff_weight}(b)), the speed weight decreases before cornering, stabilizes while navigating the corner, and gradually increases when exiting, reflecting balanced behavior between safety and efficiency.

Fig.~\ref{fig:ANC_trajectory} illustrates the generated trajectories under different driving styles in the same scenario. The aggressive style emphasizes higher speed while avoiding collisions, the normal style balances speed and collision avoidance, and the conservative style prioritizes collision avoidance at the expense of speed. This confirms that SDD Planner can accurately capture and reproduce distinct style characteristics while maintaining safety. In summary, these qualitative results demonstrate that SDD Planner not only achieves superior safety compared to baseline methods but also flexibly adapts to different driving styles, generating diverse yet safe trajectories aligned with personalized preferences.

\subsection{Ablation Study}

We conduct ablation experiments to validate the effectiveness of the proposed attention mechanism, dynamic guidance strategy, and trajectory selection rule. The analysis is organized into three parts: hyper-parameter robustness, attention/selection ablation, and full module ablation.

\subsubsection{Hyper-parameter Robustness}
To evaluate the robustness of the key hyper-parameters, we test cross-dataset transfer and parameter sweeps for $\kappa$ and $\gamma$ on NuPlan. As shown in Table~\ref{tab:robust_kappa_gamma}, directly transferring the StyleDrive-optimized $\kappa$ to NuPlan remains feasible but leads to moderate degradation, which suggests that $\kappa$ generalizes reasonably well within a stable range but still benefits from domain-specific calibration. In addition, both the $\kappa$ and $\gamma$ sweeps indicate that the default settings provide the best overall trade-off among safety, efficiency, and smoothness, validating the stability of our chosen hyper-parameters.

\begin{table}[t]
\centering
\caption{Hyper-parameter robustness on NuPlan (closed-loop). Safety, efficiency, and comfort metrics are reported under different $\kappa$/$\gamma$ settings.}
\label{tab:robust_kappa_gamma}
\renewcommand{\arraystretch}{1.05}
\scriptsize
\setlength{\tabcolsep}{3.5pt}
\begin{tabular}{lcc|ccc|cc}
\hline
\multirow{2}{*}{Setting} & \multirow{2}{*}{$\kappa$} & \multirow{2}{*}{$\gamma$} & \multicolumn{3}{c|}{Safety $\downarrow$} & Eff. $\uparrow$ & Comfort $\downarrow$ \\
\cline{4-8}
 &  &  & Coll. & TTC & Offroad & Prog. & Jerk \\
\hline
Default & 0.12 & 0.05 & 1.85 & 3.24 & 0.85 & 98.40 & 1.42 \\
Transfer $\kappa$ & 0.08 & 0.05 & 2.65 & 4.10 & 1.05 & 96.20 & 1.65 \\
$\kappa$ low & 0.05 & 0.05 & 2.15 & 5.80 & 1.10 & 91.50 & 1.88 \\
$\kappa$ mid & 0.10 & 0.05 & 1.95 & 3.55 & 0.90 & 97.80 & 1.48 \\
$\kappa$ high & 0.20 & 0.05 & 3.90 & 4.60 & 1.25 & 96.50 & 2.10 \\
$\gamma$ low & 0.12 & 0.01 & 2.45 & 4.85 & 1.15 & 94.30 & 2.35 \\
$\gamma$ high & 0.12 & 0.10 & 3.50 & 5.15 & 1.30 & 95.80 & 2.55 \\
\hline
\end{tabular}
\end{table}

\subsubsection{Attention and Selection Ablation}
To clarify the contribution of the proposed distance-sensitive attention, we compare three attention variants: (i) uniform fixed attention, where all surrounding vehicles receive equal weights; (ii) static distance weighting, where attention is determined by a fixed distance attenuation function with a non-learnable $\kappa$; and (iii) the proposed learnable distance-sensitive attention. We further compare the base trajectory output with an explicit cost-based selection strategy.

\begin{table}[t]
\centering
\caption{Ablation on attention design and trajectory selection on NuPlan.}
\label{tab:attn_selection_ablation}
\renewcommand{\arraystretch}{1.05}
\scriptsize
\setlength{\tabcolsep}{3.5pt}
\begin{tabular}{lcc|cccc}
\hline
Variant & Attn. & Select. & Coll. $\downarrow$ & TTC $\downarrow$ & Prog. $\uparrow$ & Jerk $\downarrow$ \\
\hline
Fixed-A & Uniform & Base & 6.50 & 8.20 & 89.00 & 3.10 \\
Fixed-B & Static-dist. & Base & 3.20 & 4.80 & 94.50 & 2.15 \\
Proposed & Learnable-dist. & Base & 1.85 & 3.24 & 98.40 & 1.42 \\
Proposed+ & Learnable-dist. & Cost-based & 1.70 & 3.05 & 98.60 & 1.35 \\
\hline
\end{tabular}
\end{table}

As shown in Table~\ref{tab:attn_selection_ablation}, the attention design has a substantial influence on safety and trajectory quality. Uniform fixed attention performs the worst, showing that treating all nearby agents equally is insufficient in complex interactive scenes. Introducing a static distance prior already yields clear gains, while the proposed learnable distance-sensitive attention further improves safety, progress, and smoothness. In addition, explicit cost-based candidate scoring provides further consistent gains over the base selection rule, confirming the value of trajectory set scoring for multimodal diffusion outputs.

\subsubsection{Module Ablation}
To quantify the contribution of the full framework, we further evaluate three ablated variants on NuPlan: \textbf{Variant 1 (Fixed Attention)}, which removes the adaptive distance-sensitive attention mechanism; \textbf{Variant 2 (Fixed Guidance)}, which fixes the collision and speed weights to 0.6 and 0.4; and \textbf{Variant 3 (Full Ablation)}, which removes both dynamic attention and dynamic guidance.

\begin{table}[t]
\centering
\caption{Full-module ablation on NuPlan.}
\label{tab:ablation_results}
\renewcommand{\arraystretch}{1.1}
\scriptsize
\setlength{\tabcolsep}{6pt}
\begin{tabular}{lc}
\hline
\textbf{Model} & \textbf{Total Score} \\
\hline
SDD Planner (Ours) & 91.83 \\
Variant 1 (Fixed Attention) & 85.01 \\
Variant 2 (Fixed Guidance) & 80.70 \\
Variant 3 (Full Ablation) & 77.89 \\
\hline
\end{tabular}
\end{table}

\begin{figure}
    \centering
    \includegraphics[width=1\linewidth]{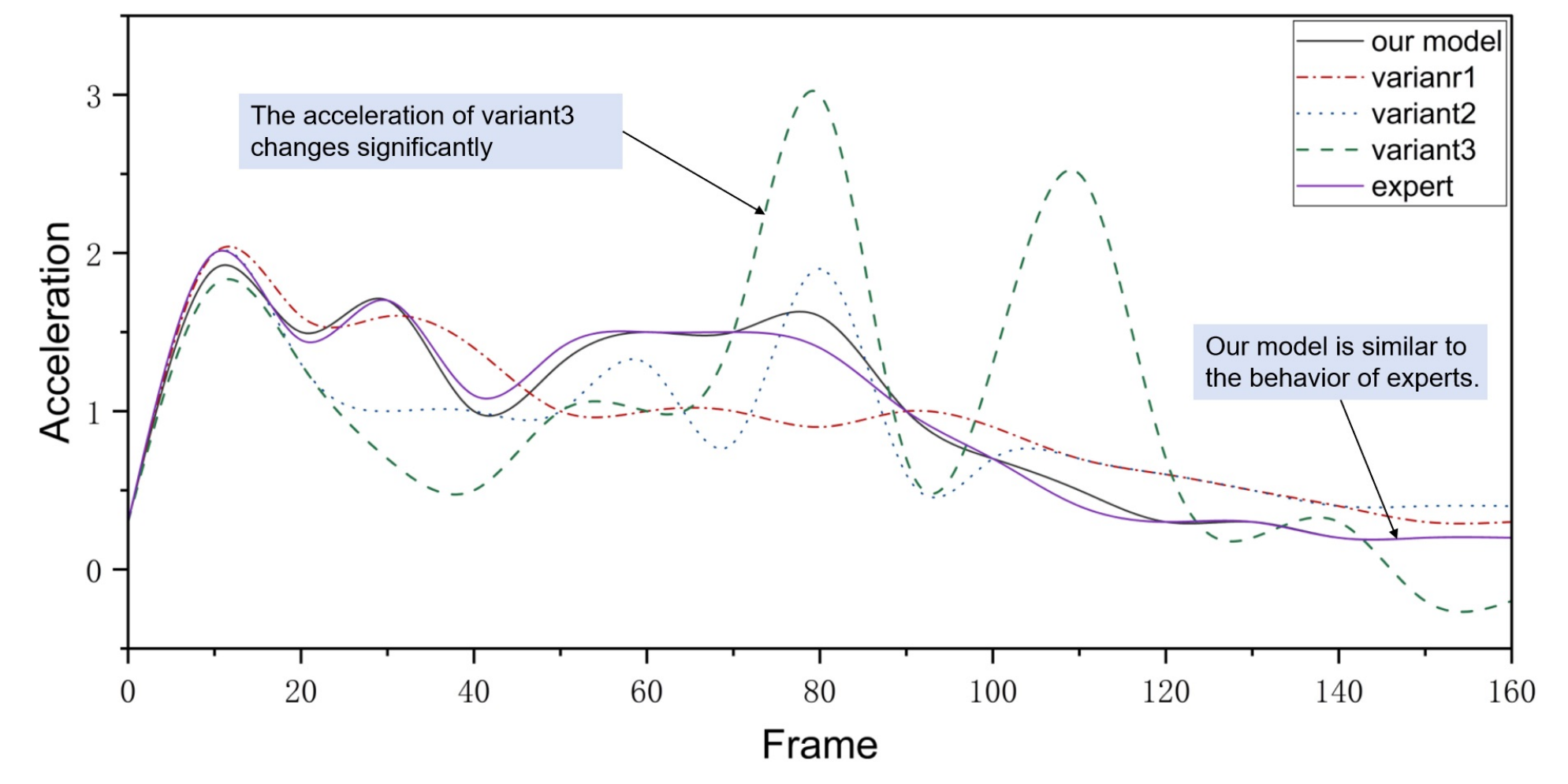}
    \caption{The acceleration of ego vehicle in the same scenario}
    \label{fig:ego_acceleration}
\end{figure}

As shown in Table~\ref{tab:ablation_results}, the full SDD Planner achieves the best overall performance, while all ablated variants show substantial degradation. As further illustrated in Fig.~\ref{fig:ego_acceleration}, the acceleration profile of our full model is closer to expert behavior and changes more smoothly than the ablated variants. In particular, Variant 3 exhibits severe acceleration fluctuations, indicating that both dynamic attention and dynamic guidance are important for safe, smooth, and controllable trajectory generation.

\subsubsection{Weights Ablation}
\begin{figure}[h]
    \centering
    \includegraphics[width=1\linewidth]{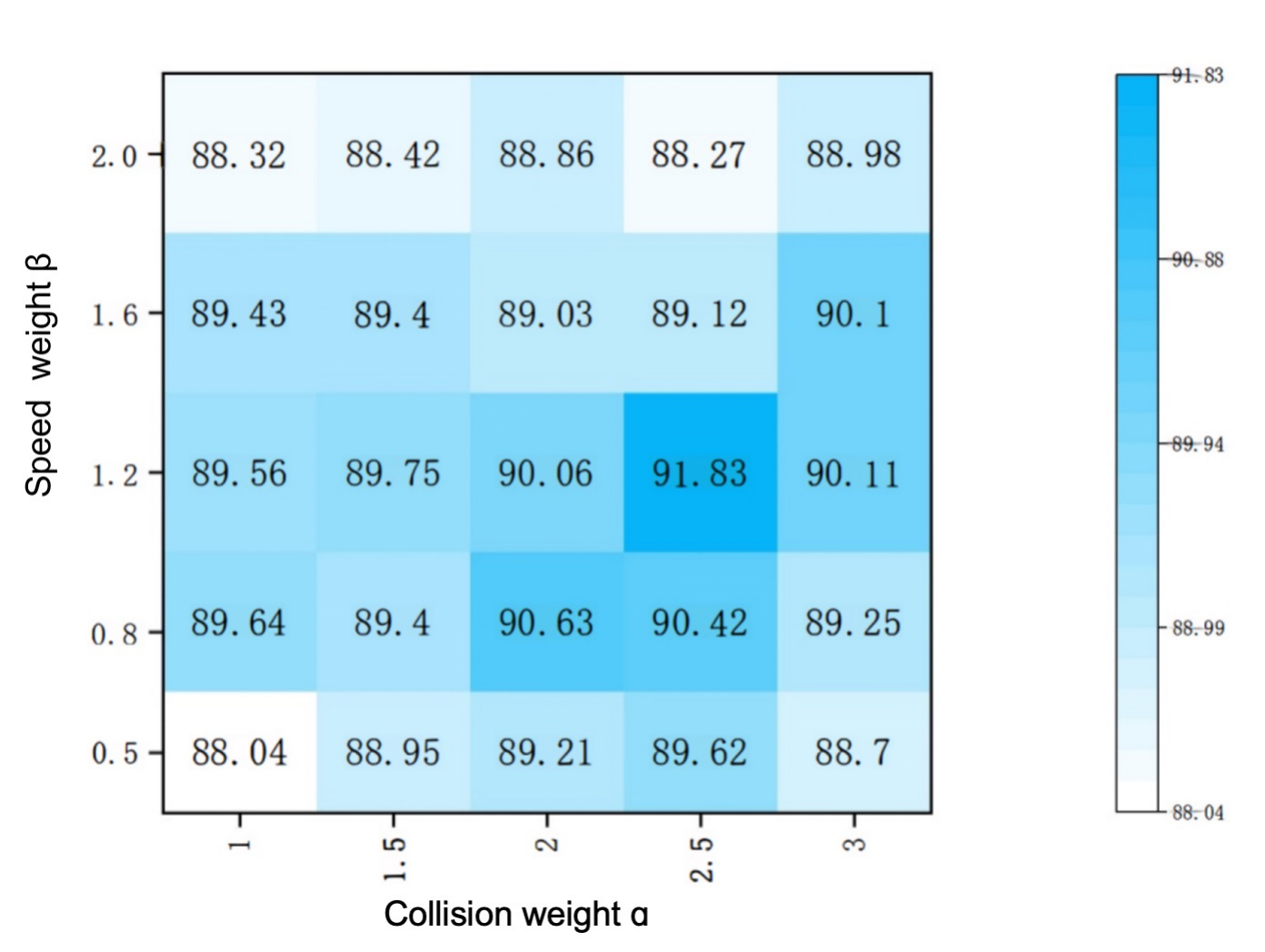}
    \caption{Performance heatmap of maximum safety weight (\(\alpha_{\text{max}}\)) and maximum speed weight (\(\beta_{\text{max}}\)). Peak performance occurs at \(\alpha_{\text{max}} = 1.2\) and \(\beta_{\text{max}} = 2.5\).}
    \label{fig:weight_heatmap}
\end{figure}

The heatmap in Fig.~\ref{fig:weight_heatmap} demonstrates that properly tuned dynamic guidance weights effectively maximize performance. The effect of maximum collision weight (\(\alpha_{\text{max}}\)) and maximum speed weight (\(\beta_{\text{max}}\)) on performance is further analyzed: (1) The values of $\alpha_{\text{max}}$ below 1.0 lead to insufficient safety regulation; values above 1.4 overly constrain trajectories, and the optimal $\alpha_{\text{max}}$ is 1.2. (2) The values of $\beta_{\text{max}}$ below 2.0 reduce speed efficiency; values above 2.8 may cause speed overload, and the optimal $\beta_{\text{max}}$ is 2.5.

Both module-level and weight-level ablations demonstrate that the dynamic attention and guidance mechanisms are essential for generating safe, comfortable, and efficient trajectories. Proper tuning of \(\alpha_{\text{max}}\) and \(\beta_{\text{max}}\) ensures peak performance of the SDD Planner.

\subsection{Real-Vehicle Deployment and Verification}
To examine the deployment feasibility of the SDD Planner, we deployed a distilled variant on an experimental vehicle equipped with an NVIDIA Orin computing platform.

\subsubsection{Model Optimization and Inference Latency}
To balance model complexity and real-time execution constraints, we applied model compression and accelerated sampling techniques. First, we utilized model distillation and post-training quantization (PTQ). The diffusion denoising pipeline was accelerated using INT8 precision, while sensitive feature extraction modules were executed in FP16. Second, the standard 1000-step diffusion process was replaced by DDIM fast sampling with 15 inference steps, significantly reducing computational overhead while preserving the overall trajectory generation behavior. Finally, it is essential to distinguish the high-level planner update period from the low-level chassis control cycle (typically $\leq 20\,\text{ms}$). The optimized SDD Planner operates at approximately 10--11 Hz, generating reference trajectories with an end-to-end latency of about $93\,\text{ms}$ \cite{wang2025modelfreecontrol}. As shown in Table~\ref{tab:vehicle_latency}, the IPC processing time comprises input preprocessing, style-aware encoding, DDIM-15 denoising, and trajectory selection.

\begin{table}[t]
\centering
\caption{Runtime profiling on the real-vehicle Orin platform.}
\label{tab:vehicle_latency}
\renewcommand{\arraystretch}{1.05}
\scriptsize
\setlength{\tabcolsep}{4pt}
\begin{tabular}{lc}
\hline
Stage & Latency (ms) \\
\hline
Input \& preprocessing & 4.5 \\
Style-aware encoder & 14.0 \\
Diffusion denoising (DDIM-15) & 66.5 \\
Selection \& output & 8.0 \\
\hline
Total & $93.0 \pm 3.0$ \\
\hline
\end{tabular}
\end{table}

\subsubsection{Closed-Loop Executability}
We evaluated the deployed planner in two representative urban scenarios to qualitatively examine its physical executability and style consistency.

\begin{figure*}[h]
    \centering
    \includegraphics[width=1\linewidth]{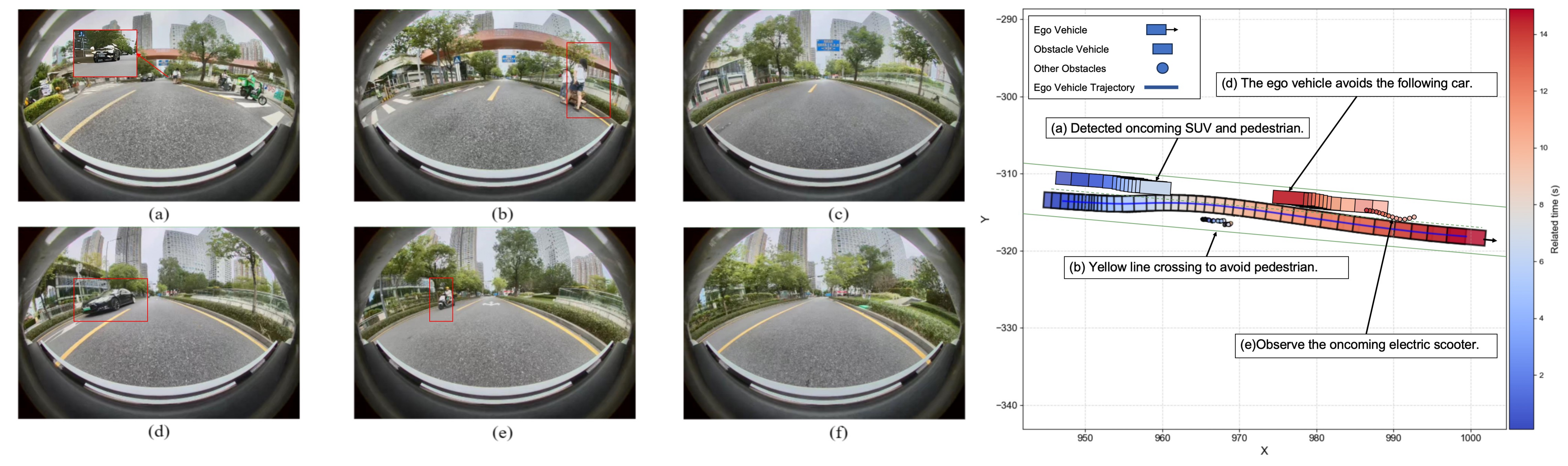}
    \caption{\textbf{Dynamic Obstacle Avoidance Scenario.} (a)--(f) Key frames from the front-facing camera showing the ego vehicle yielding to an SUV and borrowing the opposite lane to bypass a pedestrian. \textbf{Right:} Spatio-temporal trajectory color-coded by time. The visualization illustrates a smooth and cautious bypass behavior produced by the deployed planner in the experimental scenario.}
    \label{fig:rx}
\end{figure*}

\textbf{Conservative Obstacle Avoidance:} In a narrow two-way scenario, the ego vehicle was tasked with bypassing a pedestrian while interacting with oncoming traffic \cite{tang2024utdptp}. The deployed planner generated a cautious yielding strategy and then performed a controlled lane-borrowing maneuver with smooth longitudinal and lateral motion. After the lane became clear, the vehicle gradually returned to its nominal driving state. As shown in Fig.~\ref{fig:rx}, the resulting trajectory reflects a conservative interaction pattern in the presence of both vulnerable road users and oncoming vehicles.

\begin{figure*}[h]
    \centering
    \includegraphics[width=1\linewidth]{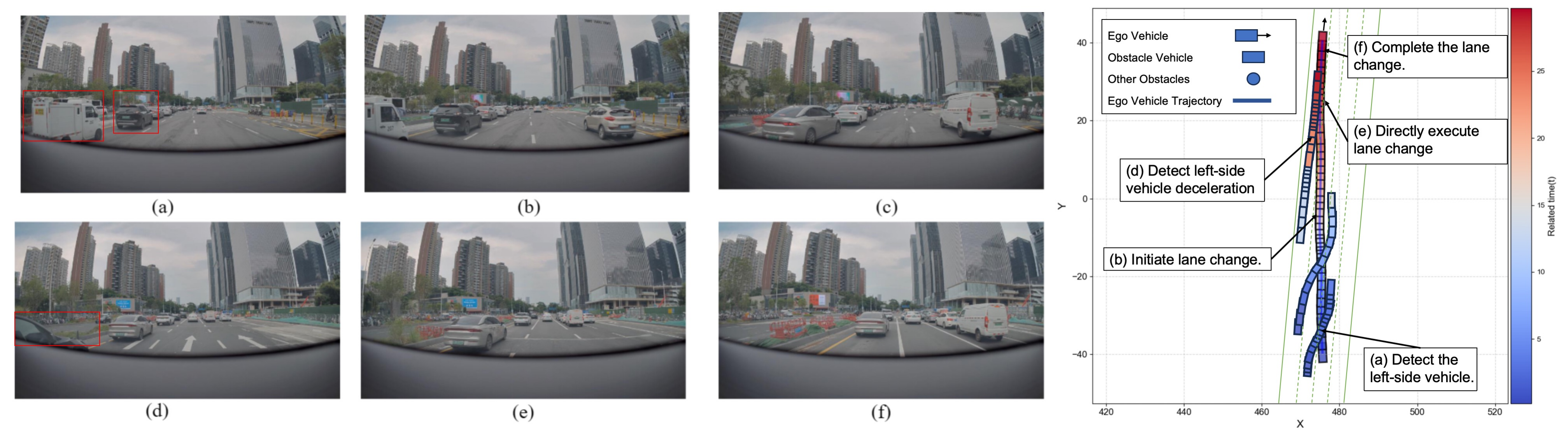}
    \caption{\textbf{Aggressive Continuous Lane Change Scenario.} (a)--(f) Front-facing camera frames showing the ego vehicle accelerating and executing a double lane change to merge in front of a yielding vehicle. \textbf{Right:} The spatio-temporal trajectory color-coded by time. The visualization highlights a more direct lateral maneuvering pattern and continuous path planning under the aggressive-style setting.}
    \label{fig:bd}
\end{figure*}

\textbf{Aggressive Lane Merging:} In an intersection scenario requiring the vehicle to merge into a far-left turning lane, the aggressive style actively sought viable gaps \cite{zhu2025hicr,tang2025ftp}. The deployed planner generated a continuous double lane-change trajectory with reduced hesitation, showing a more direct and efficiency-oriented merging behavior than the conservative setting. As illustrated in Fig.~\ref{fig:bd}, the resulting trajectory preserves the intended aggressive style tendency while remaining executable on the experimental platform.

These representative cases provide qualitative evidence that the distilled deployment variant can retain style-dependent trajectory behaviors after compression and acceleration, while remaining executable on the experimental platform. In addition, the generated trajectories are further refined before being sent to the low-level chassis controller, which helps improve kinematic feasibility during closed-loop execution \cite{li2025spatio}.

\section{CONCLUSION}
\label{sec:conclusion}

This paper presented SDD Planner, a diffusion-based trajectory planning framework for autonomous driving that explicitly reconciles safety constraints with personalized driving styles. The proposed method combines a Multi-Source Style-Aware Encoder for heterogeneous spatio-temporal context fusion and a Style-Guided Dynamic Trajectory Generator for adaptive safety--style guidance during denoising, enabling dynamic safety--style balancing and the generation of trajectories that are both safety-aware and stylistically consistent.

Extensive experiments demonstrated the effectiveness of the proposed framework. On the StyleDrive benchmark, SDD Planner achieved a state-of-the-art SM-PDMS score of 84.23, outperforming the strongest baseline by 3.9\%. On the NuPlan benchmark, it ranked first on both the Test14 and Test14-hard splits, while also improving key safety-related metrics such as collision and TTC performance. Additional robustness and ablation studies further confirmed the stability of the proposed hyper-parameter design and the effectiveness of the learnable distance-sensitive attention and cost-based trajectory selection.

A deployment-oriented study on an experimental vehicle platform further indicated the practical executability of the proposed method after model distillation, quantization, and accelerated DDIM sampling. Overall, SDD Planner provides a unified framework for user-centric autonomous driving and shows promising deployment potential under onboard runtime constraints. Future work will focus on incorporating richer semantic priors, potentially including large language models (LLMs), for style understanding, alongside exploring stronger cross-city generalization and more comprehensive system-level real-vehicle evaluation.

\bibliographystyle{IEEEtran}
\bibliography{references}

@article{li2025spatio,
  title   = {Spatio-Temporal Joint Trajectory Planning for Autonomous Vehicles Based on Improved Constrained Iterative LQR},
  author  = {Li, Qin and He, Hongwen and Hu, Manjiang and Wang, Yong},
  journal = {Sensors},
  volume  = {25},
  number  = {2},
  pages   = {512},
  year    = {2025},
  doi     = {10.3390/s25020512},
  url     = {https://www.mdpi.com/1424-8220/25/2/512}
}

@inproceedings{zheng2025diffusion,
  title     = {Diffusion-Based Planning for Autonomous Driving with Flexible Guidance},
  author    = {Zheng, Yinan and Liang, Ruiming and Zheng, Kexin and Zheng, Jinliang and Mao, Liyuan and Li, Jianxiong and Gu, Weihao and Ai, Rui and Li, Shengbo Eben and Zhan, Xianyuan and Liu, Jingjing},
  booktitle = {International Conference on Learning Representations (ICLR)},
  year      = {2025},
  doi       = {10.48550/arXiv.2501.15564},
  url       = {https://arxiv.org/abs/2501.15564}
}

@inproceedings{huang2023gameformer,
  title     = {GameFormer: Game-theoretic Modeling and Learning of Transformer-based Interactive Prediction and Planning for Autonomous Driving},
  author    = {Huang, Zhiyu and Liu, Haochen and Lv, Chen},
  booktitle = {Proceedings of the IEEE/CVF International Conference on Computer Vision (ICCV)},
  year      = {2023},
  doi       = {10.48550/arXiv.2303.05760},
  url       = {https://arxiv.org/abs/2303.05760}
}

@article{cheng2023plantf,
  title   = {Rethinking Imitation-based Planner for Autonomous Driving},
  author  = {Cheng, Jie and Chen, Yingbing and Mei, Xiaodong and Yang, Bowen and Li, Bo and Liu, Ming},
  journal = {arXiv preprint arXiv:2309.10443},
  year    = {2023},
  doi     = {10.48550/arXiv.2309.10443},
  url     = {https://arxiv.org/abs/2309.10443}
}

@article{scheel2021urbandriver,
  title   = {Urban Driving with Conditional Imitation Learning},
  author  = {Hawke, Jeffrey and Shen, Richard and Gurau, Corina and Sharma, Siddharth and Reda, Daniele and Nikolov, Nikolay and Mazur, Przemyslaw and Micklethwaite, Sean and Griffiths, Nicolas and Shah, Amar and Kendall, Alex},
  journal = {arXiv preprint arXiv:1912.00177},
  year    = {2019},
  doi     = {10.48550/arXiv.1912.00177},
  url     = {https://arxiv.org/abs/1912.00177}
}

@inproceedings{chi2023diffusion,
  title     = {Diffusion Policy: Visuomotor Policy Learning via Action Diffusion},
  author    = {Chi, Cheng and Xu, Zhenjia and Feng, Siyuan and Cousineau, Eric and Du, Yilun and Burchfiel, Benjamin and Tedrake, Russ and Song, Shuran},
  booktitle = {Robotics: Science and Systems (RSS)},
  year      = {2023},
  doi       = {10.48550/arXiv.2303.04137},
  url       = {https://arxiv.org/abs/2303.04137}
}

@inproceedings{chitta2022transfuser,
  title     = {Multi-Modal Fusion Transformer for End-to-End Autonomous Driving},
  author    = {Prakash, Aditya and Chitta, Kashyap and Geiger, Andreas},
  booktitle = {Proceedings of the IEEE/CVF Conference on Computer Vision and Pattern Recognition (CVPR)},
  pages     = {7077--7087},
  year      = {2021},
  doi       = {10.48550/arXiv.2104.09224},
  url       = {https://openaccess.thecvf.com/content/CVPR2021/html/Prakash_Multi-Modal_Fusion_Transformer_for_End-to-End_Autonomous_Driving_CVPR_2021_paper.html}
}

@article{caesar2023nuplan,
  title   = {nuPlan: A Closed-loop ML-based Planning Benchmark for Autonomous Vehicles},
  author  = {Caesar, Holger and Kabzan, Juraj and Tan, Kok Seang and Fong, Whye Kit and Wolff, Eric and Lang, Alex and Fletcher, Luke and Beijbom, Oscar and Omari, Sammy},
  journal = {arXiv preprint arXiv:2106.11810},
  year    = {2021},
  doi     = {10.48550/arXiv.2106.11810},
  url     = {https://arxiv.org/abs/2106.11810}
}

@inproceedings{dauner2023pdm,
  title     = {Parting with Misconceptions about Learning-based Vehicle Motion Planning},
  author    = {Dauner, Daniel and Hallgarten, Marcel and Geiger, Andreas and Chitta, Kashyap},
  booktitle = {Proceedings of The 7th Conference on Robot Learning},
  year      = {2023},
  doi       = {10.48550/arXiv.2306.07962},
  url       = {https://arxiv.org/abs/2306.07962}
}

@inproceedings{yang2024diffusiones,
  title     = {Diffusion-ES: Gradient-free Planning with Diffusion for Autonomous Driving and Zero-Shot Instruction Following},
  author    = {Yang, Brian and Su, Huangyuan and Gkanatsios, Nikolaos and Ke, Tsung-Wei and Jain, Ayush and Schneider, Jeff and Fragkiadaki, Katerina},
  booktitle = {Proceedings of the IEEE/CVF Conference on Computer Vision and Pattern Recognition (CVPR)},
  year      = {2024},
  doi       = {10.48550/arXiv.2402.06559},
  url       = {https://arxiv.org/abs/2402.06559}
}

@article{treiber2000intelligent,
  title   = {The Intelligent Driver Model: A Simple Car-Following Model for Highway Traffic},
  author  = {Treiber, Martin and Hennecke, Ansgar and Helbing, Dirk},
  journal = {Transportation Research Part C: Emerging Technologies},
  volume  = {8},
  number  = {5},
  pages   = {381--396},
  year    = {2000},
  doi     = {10.1016/S0968-090X(00)00003-0},
  url     = {https://doi.org/10.1016/S0968-090X(00)00003-0}
}

@article{wu2024recent,
  title   = {Recent Advances in Reinforcement Learning-Based Autonomous Driving Behavior Planning: A Survey},
  author  = {Wu, Jingda and Huang, Chao and Huang, Hailong and Lv, Chen and Wang, Yuntong and Wang, Fei-Yue},
  journal = {Transportation Research Part C: Emerging Technologies},
  volume  = {164},
  pages   = {104654},
  year    = {2024},
  doi     = {10.1016/j.trc.2024.104654},
  url     = {https://doi.org/10.1016/j.trc.2024.104654}
}

@article{tang2023multi,
  title   = {Multi-modality 3D Object Detection in Autonomous Driving: A Review},
  author  = {Tang, Yingjuan and He, Hongwen and Wang, Yong and Mao, Zan and Wang, Haoyu},
  journal = {Neurocomputing},
  volume  = {553},
  pages   = {126587},
  year    = {2023},
  doi     = {10.1016/j.neucom.2023.126587},
  url     = {https://doi.org/10.1016/j.neucom.2023.126587}
}

@article{zhang2024styledrive,
  title   = {StyleDrive: Towards Driving-Style Aware Benchmarking of End-To-End Autonomous Driving},
  author  = {Hao, Ruiyang and Jing, Bowen and Yu, Haibao and Nie, Zaiqing},
  journal = {arXiv preprint arXiv:2506.23982},
  year    = {2025},
  doi     = {10.48550/arXiv.2506.23982},
  url     = {https://arxiv.org/abs/2506.23982}
}

@article{bojarski2016end,
  title   = {End to End Learning for Self-Driving Cars},
  author  = {Bojarski, Mariusz and Del Testa, Davide and Dworakowski, Daniel and Firner, Bernhard and Flepp, Beat and Goyal, Prasoon and Jackel, Lawrence D. and Monfort, Mathew and Muller, Urs and Zhang, Jiakai and Zhang, Xin and Zhao, Jake and Zieba, Karol},
  journal = {arXiv preprint arXiv:1604.07316},
  year    = {2016},
  doi     = {10.48550/arXiv.1604.07316},
  url     = {https://arxiv.org/abs/1604.07316}
}

@inproceedings{sohl2015deep,
  title     = {Deep Unsupervised Learning Using Nonequilibrium Thermodynamics},
  author    = {Sohl-Dickstein, Jascha and Weiss, Eric A. and Maheswaranathan, Niru and Ganguli, Surya},
  booktitle = {Proceedings of the 32nd International Conference on Machine Learning (ICML)},
  pages     = {2256--2265},
  year      = {2015},
  doi       = {10.48550/arXiv.1503.03585},
  url       = {https://proceedings.mlr.press/v37/sohl-dickstein15.html}
}

@article{zhai2023rethinking,
  title   = {Rethinking the Open-Loop Evaluation of End-to-End Autonomous Driving in nuScenes},
  author  = {Zhai, Jiang-Tian and Feng, Ze and Du, Jinhao and Mao, Yongqiang and Liu, Jiang-Jiang and Tan, Zichang and Zhang, Yifu and Ye, Xiaoqing and Wang, Jingdong},
  journal = {arXiv preprint arXiv:2305.10430},
  year    = {2023},
  doi     = {10.48550/arXiv.2305.10430},
  url     = {https://arxiv.org/abs/2305.10430}
}

@inproceedings{ho2020denoising,
  title     = {Denoising Diffusion Probabilistic Models},
  author    = {Ho, Jonathan and Jain, Ajay and Abbeel, Pieter},
  booktitle = {Advances in Neural Information Processing Systems (NeurIPS)},
  volume    = {33},
  pages     = {6840--6851},
  year      = {2020},
  doi       = {10.48550/arXiv.2006.11239},
  url       = {https://arxiv.org/abs/2006.11239}
}

@inproceedings{janner2022diffuser,
  title     = {Planning with Diffusion for Flexible Behavior Synthesis},
  author    = {Janner, Michael and Du, Yilun and Tenenbaum, Joshua B. and Levine, Sergey},
  booktitle = {International Conference on Machine Learning (ICML)},
  pages     = {9902--9915},
  year      = {2022},
  url       = {https://proceedings.mlr.press/v162/janner22a.html}
}

@article{cheng2024pluto,
  title   = {PLUTO: Pushing the Limit of Imitation Learning-based Planning for Autonomous Driving},
  author  = {Cheng, Jie and Chen, Yingbing and Chen, Qifeng},
  journal = {arXiv preprint arXiv:2404.14327},
  year    = {2024},
  doi     = {10.48550/arXiv.2404.14327},
  url     = {https://arxiv.org/abs/2404.14327}
}

@article{dinneweth2022marlsurvey,
  title   = {Multi-agent Reinforcement Learning for Autonomous Vehicles: A Survey},
  author  = {Dinneweth, Joris and Boubezoul, Abderrahmane and Mandiau, Ren{\'e} and Espi{\'e}, St{\'e}phane},
  journal = {Autonomous Intelligent Systems},
  year    = {2022},
  volume  = {2},
  number  = {1},
  pages   = {1--27},
  doi     = {10.1007/s43684-022-00045-z},
  url     = {https://link.springer.com/article/10.1007/s43684-022-00045-z}
}

@article{zhu2025hicr,
  title   = {Multi-agent Trajectory Prediction with Hierarchical Coordinate-Based Representation},
  author  = {Zhu, Yuanchen and Fu, Shuaiqi and Wang, Yong and Zhao, Yanan and Tan, Huachun},
  journal = {Engineering Applications of Artificial Intelligence},
  year    = {2025},
  pages   = {113335},
  doi     = {10.1016/j.engappai.2025.113335},
  url     = {https://doi.org/10.1016/j.engappai.2025.113335}
}

@article{wang2025modelfreecontrol,
  title   = {Model-Free Control Framework for Stability and Path-tracking of Autonomous Independent-Drive Vehicles},
  author  = {Wang, Yong and Tang, Jianming and Li, Qin and Zhao, Yanan and Sun, Chen and He, Hongwen},
  journal = {IEEE Transactions on Transportation Electrification},
  year    = {2025},
  doi     = {10.1109/TTE.2025.3563395},
  url     = {https://doi.org/10.1109/TTE.2025.3563395}
}

@article{tang2024utdptp,
  title   = {Utilizing a Diffusion Model for Pedestrian Trajectory Prediction in Semi-Open Autonomous Driving Environments},
  author  = {Tang, Yingjuan and He, Hongwen and Wang, Yong and Wu, Yifan},
  journal = {IEEE Sensors Journal},
  year    = {2024},
  doi     = {10.1109/JSEN.2024.3382406},
  url     = {https://doi.org/10.1109/JSEN.2024.3382406}
}

@article{tang2025ftp,
  title   = {Flexible Anchor-Based Trajectory Prediction for Different Types of Traffic Participants in Autonomous Driving Systems},
  author  = {Tang, Yingjuan and He, Hongwen and Wang, Yong and Wu, Yifan},
  journal = {Expert Systems with Applications},
  volume  = {282},
  pages   = {127629},
  year    = {2025},
  doi     = {10.1016/j.eswa.2025.127629},
  url     = {https://doi.org/10.1016/j.eswa.2025.127629}
}

@inproceedings{sima2024drivelm,
  title     = {DriveLM: Driving with Graph Visual Question Answering},
  author    = {Sima, Chonghao and Renz, Katrin and Chitta, Kashyap and Chen, Li and Zhang, Hanxue and Xie, Chengen and Beisswenger, Jens and Luo, Ping and Geiger, Andreas and Li, Hongyang},
  booktitle = {European Conference on Computer Vision (ECCV)},
  year      = {2024},
  doi       = {10.1007/978-3-031-72673-6_13},
  url       = {https://arxiv.org/abs/2312.14150}
}

@article{shao2023lmdrive,
  title   = {LMDrive: Closed-Loop End-to-End Driving with Large Language Models},
  author  = {Shao, Hao and Hu, Yuxuan and Wang, Letian and Waslander, Steven L. and Liu, Yu and Li, Hongsheng},
  journal = {arXiv preprint arXiv:2312.07488},
  year    = {2023},
  doi     = {10.48550/arXiv.2312.07488},
  url     = {https://arxiv.org/abs/2312.07488}
}

@article{li2025wote,
  title   = {End-to-End Driving with Online Trajectory Evaluation via BEV World Model},
  author  = {Li, Yingyan and Wang, Yuqi and Liu, Yang and He, Jiawei and Fan, Lue and Zhang, Zhaoxiang},
  journal = {arXiv preprint arXiv:2504.01941},
  year    = {2025},
  doi     = {10.48550/arXiv.2504.01941},
  url     = {https://arxiv.org/abs/2504.01941}
}
\end{document}